\documentclass[journal]{IEEEtran}

\IEEEoverridecommandlockouts    % Needed for \thanks command
\newcommand\revise[1]{\textcolor{black}{#1}}
\newcommand\todo[1]{\textcolor{black}{#1}}

\newcommand\lz[1]{\textcolor{black}{#1}}

\newcommand{\boldcheckmark}{\ding{52}}

\usepackage[font=small]{caption}

\usepackage{glossaries}

\newacronym{mav}{UAV}{Unmanned Aerial Vehicle}
\newacronym{uav}{UAV}{Unmanned Aerial Vehicle}
\newacronym{ovc}{OVC}{Open Vision Computer}
\newacronym{lidar}{LiDAR}{Light Detection and Ranging}
\newacronym{vio}{VIO}{visual-inertial odometry}
\newacronym{satnav}{GNSS}{Global Navigation Satellite Systems}
\newacronym{dgps}{DGPS}{Differential GPS}
\newacronym{rtk}{RTK}{Real-Time Kinematics}
\newacronym{ppk}{PPK}{Post-Processed Kinematic}
\newacronym{gpgpu}{GPGPU}{General-Purpose Graphics Processing Unit}
\newacronym{hri}{HRI}{Human-Robot Interaction}
\newacronym{ugv}{UGV}{Unmanned Ground Vehicle}
\newacronym{uwb}{UWB}{Ultra Wideband}
\newacronym{svm}{SVM}{Support Vector Machine}
\newacronym{fcn}{FCN}{Fully Convolutional Network}
\newacronym{cnn}{CNN}{Convolutional Neural Network}
\newacronym{loam}{LOAM}{LiDAR Odometry and Mapping}
\newacronym{sloam}{SLOAM}{Semantic LiDAR Odometry and Mapping}
\newacronym{slam}{SLAM}{Simultaneous Localization and Mapping}
\newacronym{swap}{SWaP}{Size, Weight, and Power}
\newacronym{iot4ag}{IoT4Ag}{NSF Engineering Research Center for the Internet of Things for Precision Agriculture}
\newacronym{grasp-lab}{GRASP Lab}{the General Robotics, Automation, Sensing and Perception Laboratory}
\newacronym{jps}{JPS}{Jump Point Search}
\newacronym{ukf}{UKF}{Unscented Kalman Filter}
\newacronym{sam}{SAM}{Smoothing and Mapping}
\newacronym{icp}{ICP}{Iterative Closest Point}
\newacronym{imu}{IMU}{Inertial Measurement Unit}
\newacronym{tsdf}{TSDF}{Truncated Signed Distance Field}
\newacronym{esdf}{ESDF}{Euclidean Signed Distance Field}
\newacronym{rrt}{RRT}{A rapidly exploring random tree}
\usepackage[table,usenames,dvipsnames]{xcolor}      % color
\usepackage{extarrows}                              % http://ctan.org/pkg/extarrows
\usepackage{enumitem}
\usepackage{wrapfig}

\usepackage{tikz}
\usepackage{varwidth}
\usepackage{multirow, multicol}
\usepackage{array}
\newcolumntype{P}[1]{>{\centering\arraybackslash}p{#1}}
\newcolumntype{M}[1]{>{\centering\arraybackslash}m{#1}}
\usepackage{booktabs}
\newcolumntype{N}{>{\centering\arraybackslash}m{.5in}}
\newcolumntype{G}{>{\centering\arraybackslash}m{2in}}

\usepackage{amsmath,amssymb,amsfonts,amsthm,dsfont} % math
\usepackage{algorithm, algorithmicx, listings}        % algorithms
\usepackage[noend]{algpseudocode}			              % necessary for algorithmicx
\makeatletter
\def\BState{\State\hskip-\ALG@thistlm}
\makeatother

\usepackage{graphicx}
\usepackage{tabularx}
\usepackage{subcaption}
\usepackage{textcomp}

\usepackage[breaklinks=true, colorlinks, bookmarks=true, citecolor=Black, urlcolor=Violet,linkcolor=Black]{hyperref}

\usepackage{svg}

\usepackage{amssymb}

\def\deg{^{\circ}}
\newcolumntype{C}[1]{>{\centering\arraybackslash}p{#1}}

\usepackage{colortbl}
\definecolor{dark-grey}{rgb}{0.8, 0.8, 0.8}
\definecolor{original}{rgb}{0, 0, 0}  % Default black color

\usepackage{everypage}
\usepackage{afterpage}
\usepackage{ifthen}

\usepackage[normalem]{ulem}
\usepackage[utf8]{inputenc}
\usepackage[english]{babel}
\usepackage{hyperref}
\hypersetup{
    colorlinks=true,
    linkcolor=blue,
    filecolor=magenta,      
    urlcolor=cyan,
}

\newcommand\copyrighttext{%
  \footnotesize © 2025 IEEE. Personal use of this material is permitted. 
  Permission from IEEE must be obtained for all other uses, in any current or future
  media, including reprinting/republishing this material for advertising or promotional
  purposes, creating new collective works, for resale or redistribution to servers or
  lists, or reuse of any copyrighted component of this work in other works. This is the author’s accepted version of an article in \textit{IEEE Transactions on Robotics (T-RO)}. 
}

\newcommand\copyrightnotice{%
  \begin{tikzpicture}[remember picture,overlay]
    \node[anchor=south,yshift=5pt] at (current page.south) {%
      \fbox{%
        \begin{varwidth}{1.0\textwidth}
          \copyrighttext
        \end{varwidth}%
      }};
  \end{tikzpicture}%
}

\DeclareMathAlphabet\mathbfcal{OMS}{cmsy}{b}{n}

\newtheorem*{assumption*}{Assumption}

\newtheorem*{problem*}{Problem}

\usepackage{lipsum}
\usepackage[capitalise]{cleveref}
\usepackage{svg}
\usepackage{mathtools}
\usepackage{pifont}

\begin{document}
\title{SlideSLAM: Sparse, Lightweight, Decentralized Metric-Semantic SLAM for Multi-Robot Navigation}

\author{Xu Liu*$^{\dagger}$, Jiuzhou Lei*, Ankit Prabhu*, Yuezhan Tao, Igor Spasojevic, \\ Pratik Chaudhari, Nikolay Atanasov, Vijay Kumar

\thanks{* Equal contribution. $^{\dagger}$ Corresponding author.}
\thanks{Manuscript received January 9, 2025; revised May 1, 2025; accepted September 2, 2025. 
This article was recommended for publication by Editor Javier Civera upon evaluation of the reviewers’ comments. 
This work was supported by funding from the IoT4Ag Engineering Research Center funded by NSF (NSF EEC-1941529), National Robotics Initiative 3.0: Innovations in Integration of Robotics (NSF 21-559), NIFA grant 2022-67021-36856, NSF grant CCR-2112665, and the ARL DCIST CRA W911NF-17-2-0181. 
}
\thanks{X. Liu is with Microsoft, Redmond, WA 98052, USA (e-mail: {\tt\small {liuxushawn@gmail.com}}). This work was done while X. Liu was with the GRASP Laboratory, University of Pennsylvania, Philadelphia, PA 19104, USA.}
\thanks{J. Lei, A. Prabhu, Y. Tao, I. Spasojevic, P. Chaudhari, and V. Kumar are with the GRASP Laboratory, University of Pennsylvania, Philadelphia, PA 19104, USA (e-mail: {\tt\small{\{jiuzl, praankit, yztao, igorspas, pratikac, kumar\}@upenn.edu}}). }
\thanks{N. Atanasov is with the Department of Electrical and Computer Engineering, University of California San Diego, La Jolla, CA 92093, USA (e-mail: {\tt\small {natanasov@ucsd.edu}}).} 
}

% Paper headers 
\markboth{IEEE Transactions on Robotics. Preprint Version. Accepted September 2025}{Liu \MakeLowercase{\textit{et al.}}: SlideSLAM}

\maketitle

\copyrightnotice

% \IEEEpubidadjcol 
% % Force copyright block across both columns
% \IEEEpubid{\makebox[\textwidth]{%
%   © 2025 IEEE. This is the author’s accepted version of an article that will appear in 
%   \textit{IEEE Transactions on Robotics (T-RO)}. 
%   The final published version is available at: https://doi.org/10.1109/TRO.2025.xxxxx. 
%   Personal use of this material is permitted. Permission from IEEE must be obtained for all 
%   other uses, in any current or future media, including reprinting/republishing this material 
%   for advertising or promotional purposes, creating new collective works, for resale or 
%   redistribution to servers or lists, or reuse of any copyrighted component of this work in 
%   other works.\hfill}}
% \IEEEpubidadjcol

\begin{abstract}
This paper develops a real-time decentralized metric-semantic SLAM algorithm that enables a heterogeneous robot team to collaboratively construct object-based metric-semantic maps. The proposed framework integrates a data-driven front-end for instance segmentation from either RGBD cameras or LiDARs and a custom back-end for optimizing robot trajectories and object landmarks in the map. To allow multiple robots to merge their information, we design semantics-driven place recognition algorithms that leverage the informativeness and viewpoint invariance of the object-level metric-semantic map for inter-robot loop closure detection. A communication module is designed to track each robot's observations and those of other robots whenever communication links are available. The framework supports real-time, decentralized operation onboard the robots and has been integrated with three types of aerial and ground platforms. We validate its effectiveness through experiments in both indoor and outdoor environments, as well as benchmarks on public datasets and comparisons with existing methods. The framework is open-sourced and suitable for both single-agent and multi-robot real-time metric-semantic SLAM applications. 
The code is available at: \url{https://github.com/KumarRobotics/SLIDE_SLAM}.

\end{abstract}

\section{Introduction}
\label{sec:introduction}

\begin{figure}[t!]
        \centering
        \includegraphics[trim=0 0 0 0, clip, width=1.0\columnwidth]{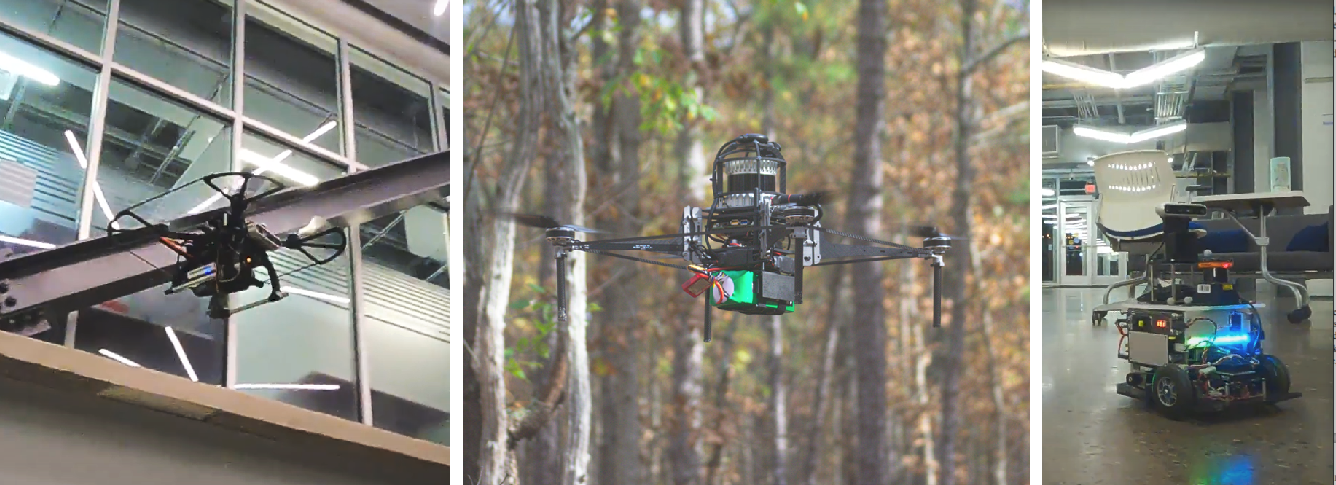}
        \caption{\textit{Robot platforms used in our experiments}. We utilize three types of robots for our experiments: two aerial platforms, the Falcon 250 UAV (left) and the Falcon 4 UAV (middle), and one ground platform, the Scarab UGV (right). The Light Detection and Ranging (LiDAR)-equipped robot (Falcon 4) is primarily used for outdoor operations due to its size and superior sensing capabilities. The RGB and Depth (RGBD) camera-based robots (Falcon 250 and Scarab) are more suitable for cluttered indoor environments due to their smaller footprints. All three platforms have GPS-denied autonomous navigation capabilities, enabling them to safely explore cluttered environments using only onboard computation and sensing.}
        % \vspace{-0.1in}
        \label{fig:title-page}
\end{figure}

\begin{figure*}[t!]
    \centering
    \begin{subfigure}[t]{0.49\textwidth}
        \centering
        \includegraphics[width=0.99\columnwidth]{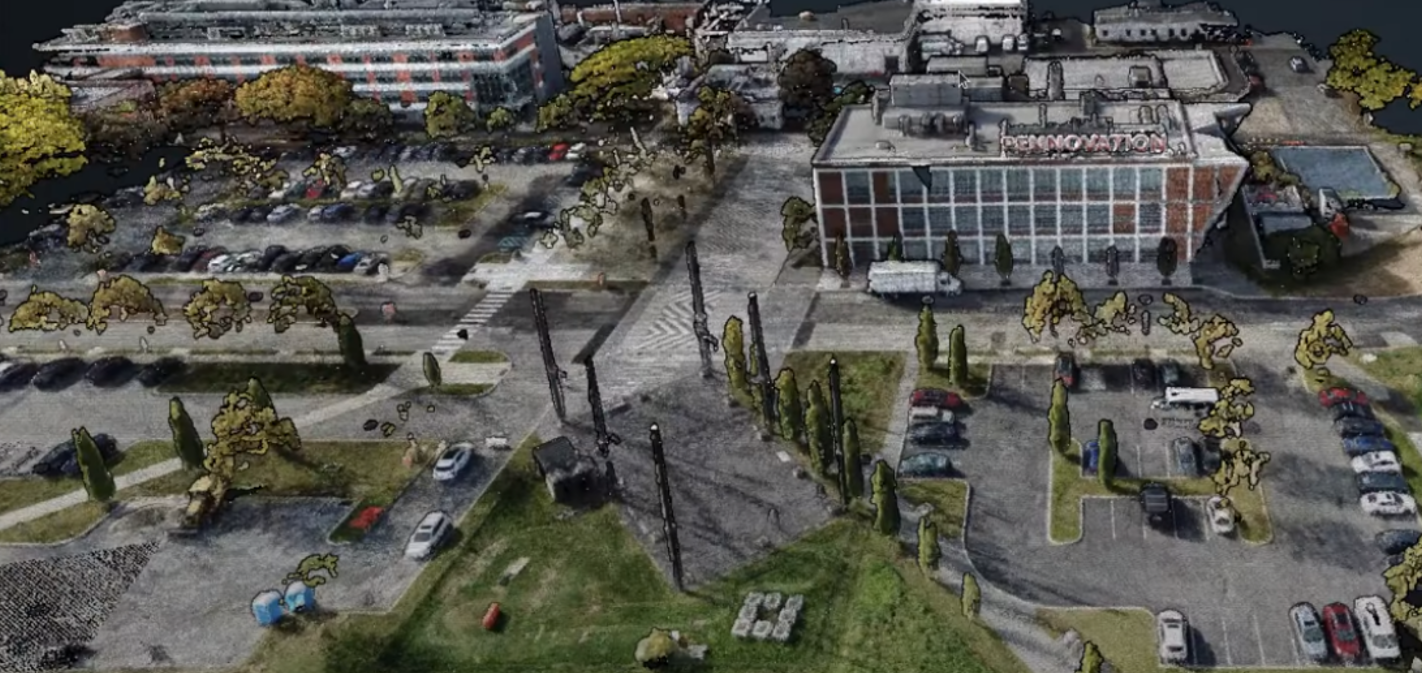}
        \caption{3D reconstruction}
        \label{fig:subfigure a}
    \end{subfigure}%
    ~%
    \begin{subfigure}[t]{0.49\textwidth}
        \centering
       \includegraphics[width=0.99\columnwidth]{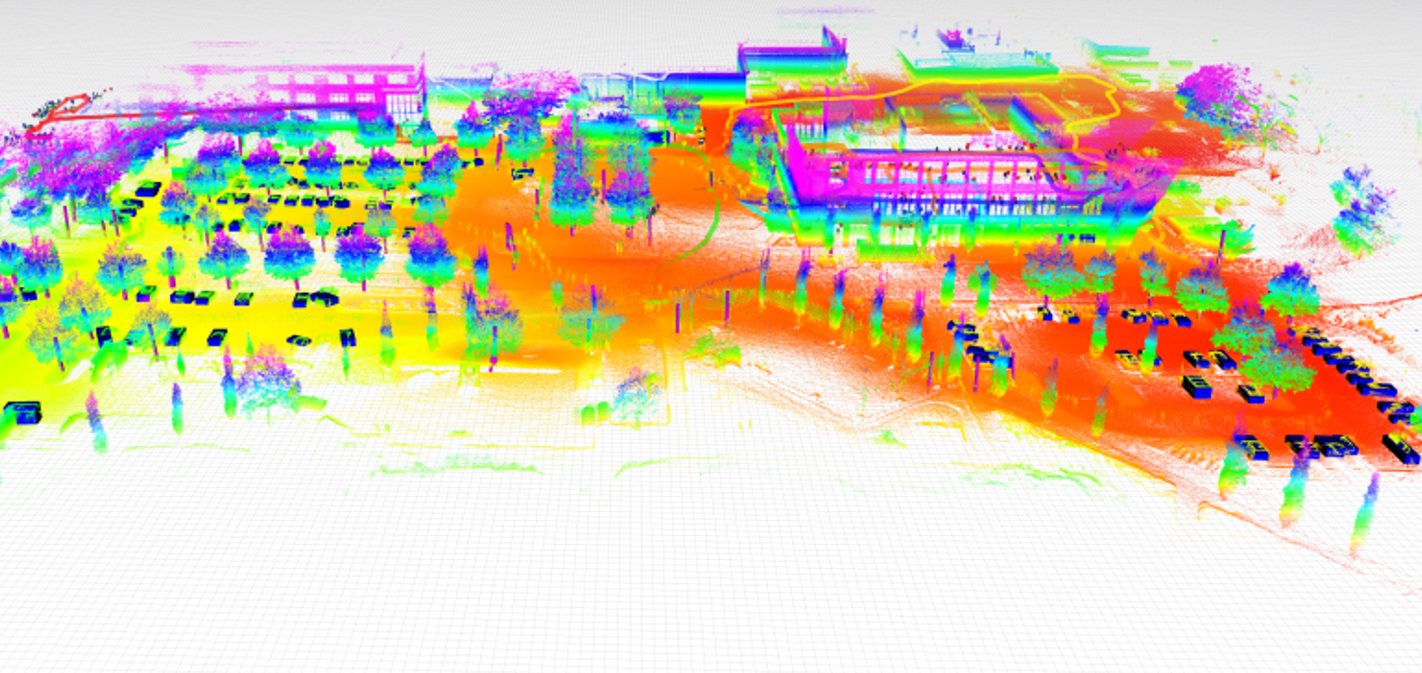}
        \caption{metric-semantic map with point cloud}
        \label{fig:subfigure b}
    \end{subfigure}
    \begin{subfigure}[t]{0.49\textwidth}
        \centering
        \includegraphics[width=0.99\columnwidth]{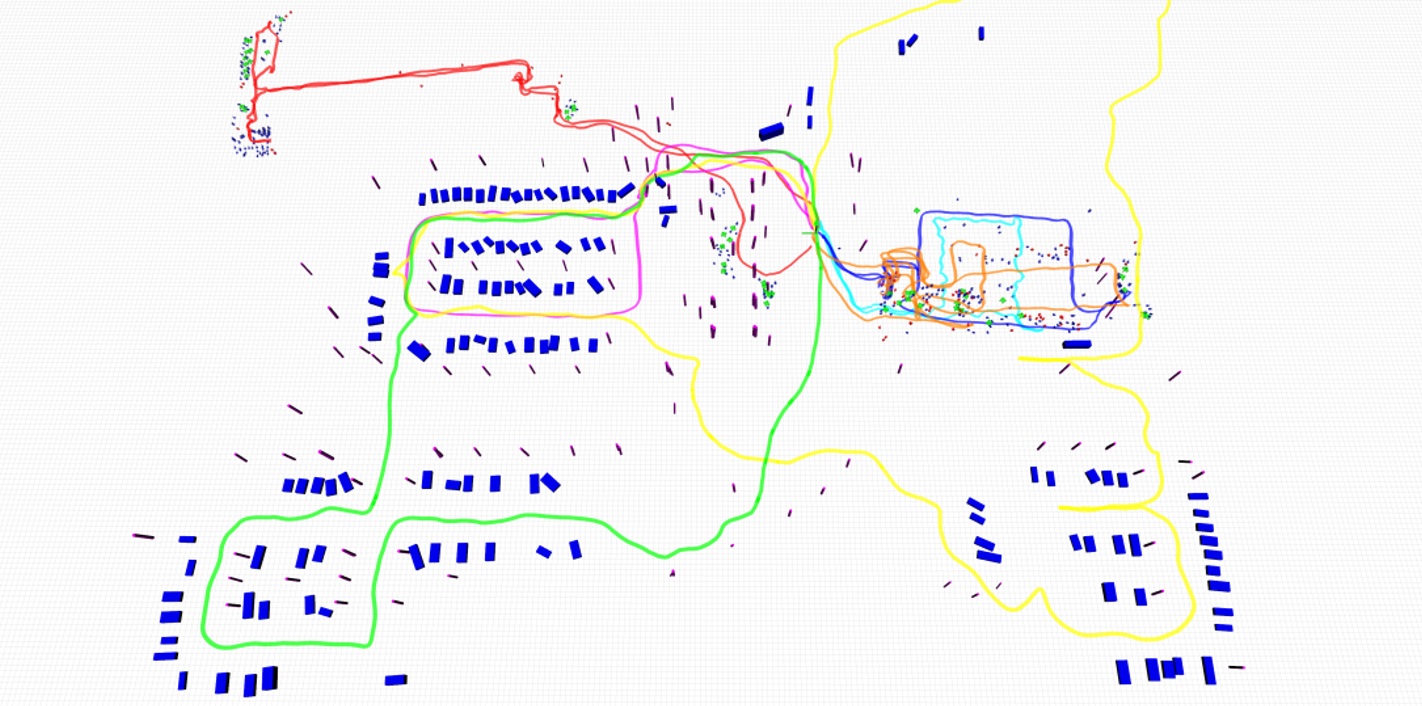}
        \caption{object-level map top view}
        \label{fig:subfigure c}
    \end{subfigure}%
    \begin{subfigure}[t]{0.49\textwidth}
        \centering
        \includegraphics[ width=0.99\columnwidth]{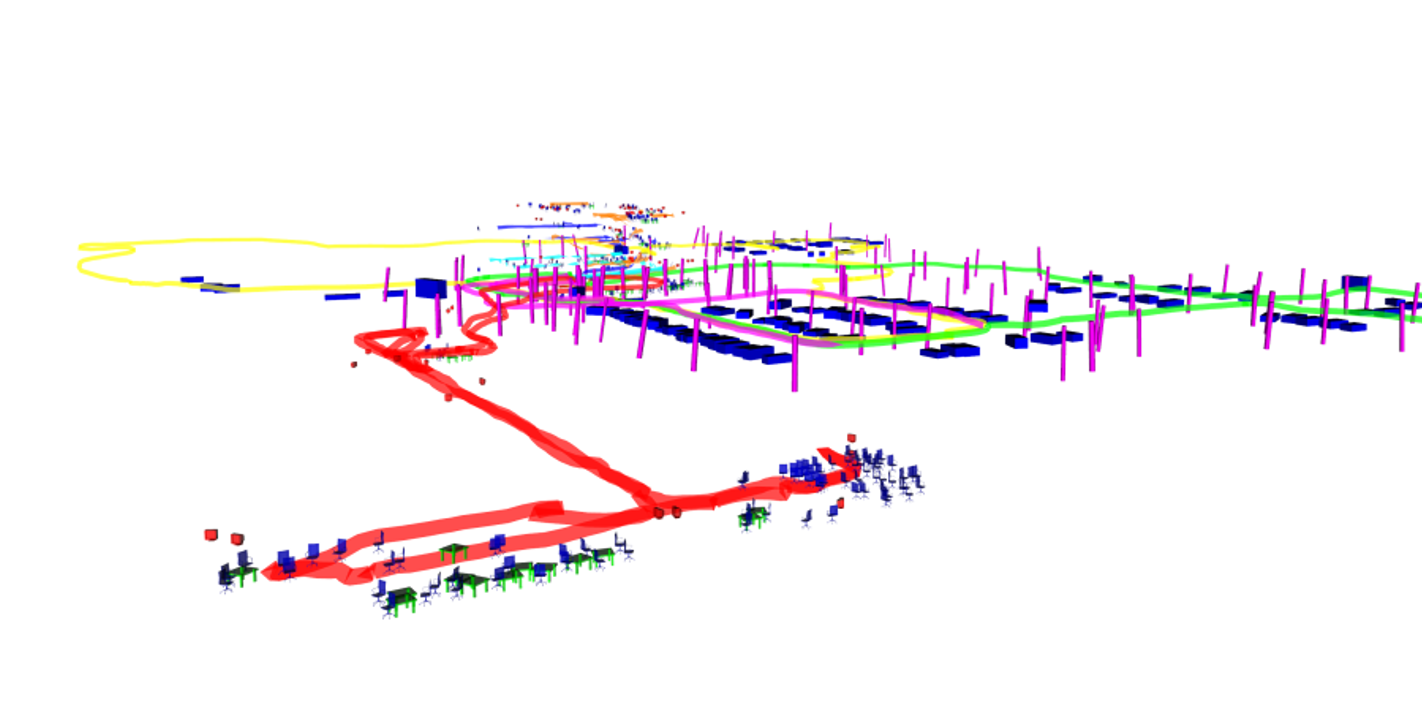}
        \caption{object-level map oblique view}
        \label{fig:subfigure d}
    \end{subfigure}
    \begin{subfigure}[t]{0.49\textwidth}
        \centering
        \includegraphics[width=0.99\columnwidth]{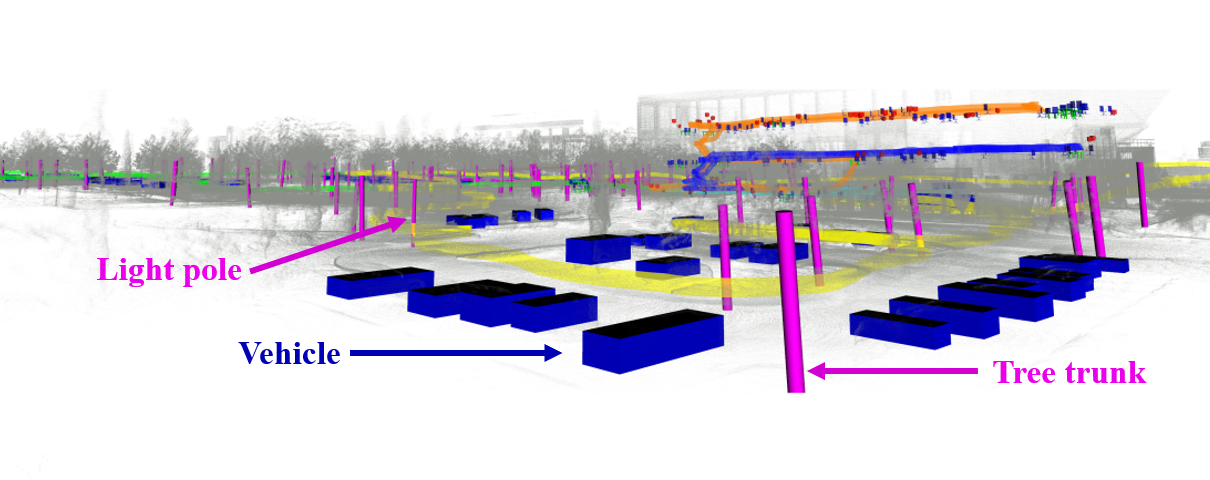}
        \caption{outdoor objects close-up view}
        \label{fig:subfigure e}
    \end{subfigure}%
    \begin{subfigure}[t]{0.49\textwidth}
        \centering
        \includegraphics[ width=0.99\columnwidth]{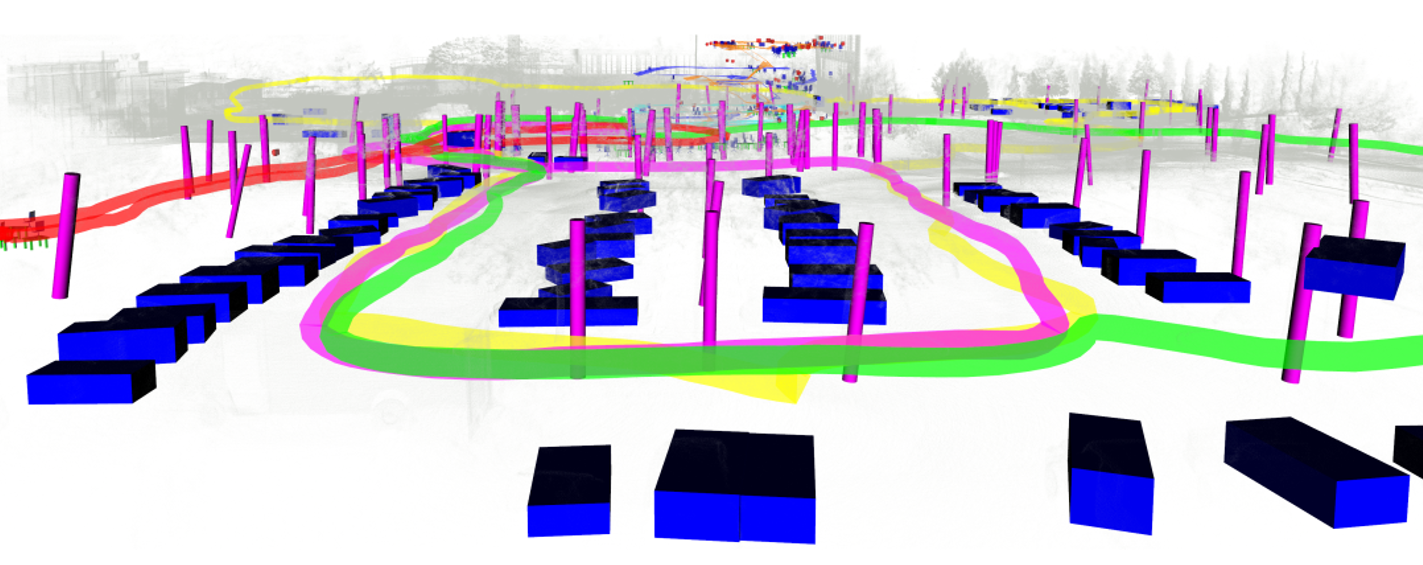}
        \caption{parking lot close-up view}
        \label{fig:subfigure f}
    \end{subfigure}
    \begin{subfigure}[t]{0.49\textwidth}
        \centering
        \includegraphics[ width=0.99\columnwidth]{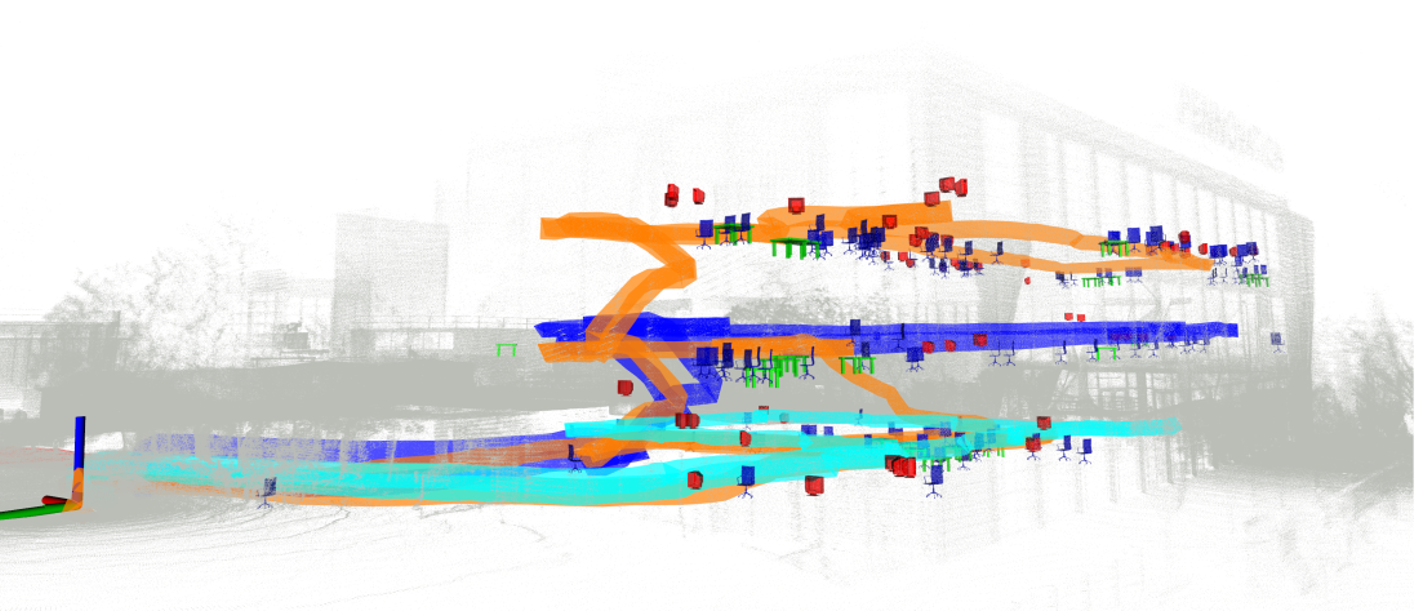}
        \caption{building close-up view}
        \label{fig:subfigure g}
    \end{subfigure}%
    \begin{subfigure}[t]{0.49\textwidth}
        \centering
        \includegraphics[width=0.99\columnwidth]{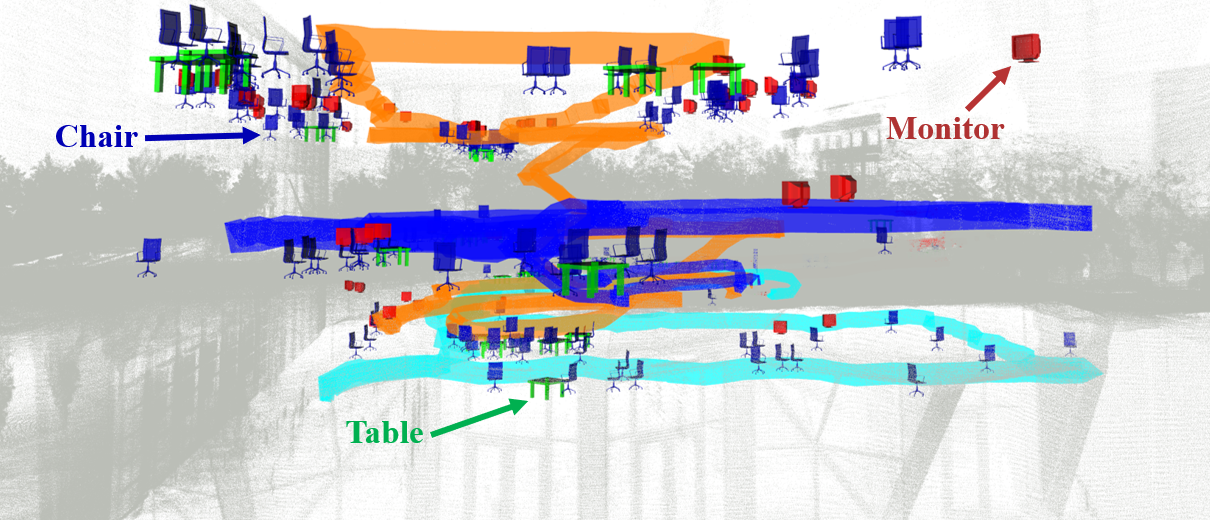}
        \caption{indoor objects close-up view}
        \label{fig:subfigure h}
    \end{subfigure}
    \caption{\textit{Metric-semantic SLAM results from seven data sequences collected by heterogeneous robots}. Trajectories in different colors correspond to different data sequences. Fig.~\ref{fig:subfigure a} shows a 3D reconstruction of the Pennovation campus at the University of Pennsylvania. 
    Outdoor objects, such as vehicles, tree trunks, and light poles are mapped as shown in \ref{fig:subfigure e}. Indoor objects, such as chairs, tables, and monitors are mapped as shown in Fig.~\ref{fig:subfigure h}. Fig.~\ref{fig:subfigure b} shows the same metric-semantic map overlayed on top of an accumulated point cloud constructed by our Falcon 4 UAV. Fig.~\ref{fig:subfigure c} shows an orthophoto depicting the merged metric-semantic map of three parking lots and two buildings constructed by seven robots. Fig.~\ref{fig:subfigure g} and Fig.~\ref{fig:subfigure h} show a zoomed-in view of one of the lab buildings}%, and illustrate the metric-semantic qualitative mapping performance in the indoor environments.}
    % \vspace{-0.1in}
\label{fig:seven-robot-indoor-outdoor}
\end{figure*}

Robotic systems are expected to make an impact in demanding applications, such as forest inventory management, orchard yield estimation, infrastructure inspection and household assistance. This demands interpreting human instructions given in semantically meaningful terms relating to objects and properties in the robot's environment. To execute such missions autonomously, robots must understand their environment beyond its geometric structure and perceive it at a semantic level. This requires building and maintaining a semantically meaningful representation of the environment that encodes actionable information (e.g., timber volume and health in forests, corrosion in infrastructure, survivors' locations in natural disasters). Such a representation has to be storage efficient for the robots to maintain over large-scale missions, and has to allow efficient optimization for Simultaneous Localization and Mapping (SLAM).

While traditional SLAM approaches \cite{orbslam3, shan2020lio} offer excellent accuracy in geometric perception, including estimating robot poses and reconstructing 3D geometric structures (points, surfaces, voxels), they are often insufficient to support large-scale missions, particularly when used with multiple heterogeneous robots for real-time autonomy. The challenges include managing the excessive computational load, meeting the high storage demands of large-scale maps, and handling loop closure and map merging operations. Moreover, they offer limited generalizability across different robot platforms or sensing modalities and lack semantic map representations required for executing semantically meaningful tasks.

\begin{table*}[t!]
\centering
\renewcommand{\arraystretch}{1.25}
     \caption{\textit{Related work.} 
    This table compares SlideSLAM against state-of-the-art methods, focusing on key attributes for enabling semantics-in-the-loop autonomy with robot teams. 
    In this table, decentralized methods support operation with asynchronous observation updates from other robots. 
    Semantic localization means that objects are used for localization between consecutive key poses.
    Semantic loop closure indicates that objects are explicitly used for place recognition and loop closures. 
    Optimization over object models involves explicitly modeling objects as geometric shapes and optimizing them jointly with robot poses over time. 
    Real-time with autonomy means that the metric-semantic SLAM system operates in real-time on-board the robots and is fully integrated into an autonomous exploration or navigation system. 
    % Open-vocabulary refers to the capability of robots to perceive open-vocabulary semantic information and map objects beyond predefined categories. 
    % We develop the first metric-semantic SLAM algorithm that supports decentralized operation and is deployed to enable autonomous navigation and exploration using heterogeneous robot teams.
    }
\resizebox{0.9\textwidth}{!}
{
\arrayrulecolor{dark-grey}
 \begin{tabular}{c |c | c | c | c | c  | c | c  }
        % \hline
                     & Year                          & Multi-Robot                   & \begin{tabular}[c]{@{}c@{}}Decentralized, \\ Asynchronous\end{tabular} & \multicolumn{1}{c|}{\begin{tabular}[c]{@{}c@{}}Semantic \\ Localization\end{tabular}} & \multicolumn{1}{c|}{\begin{tabular}[c]{@{}c@{}}Semantic\\ Loop Closure\end{tabular}} & \multicolumn{1}{c|}{\begin{tabular}[c]{@{}c@{}}Optimization over\\ Object Models\end{tabular}} & {\begin{tabular}[c]{@{}c@{}}Real-time w/ \\  Autonomy \end{tabular}}  \\ 
                     \hline
        \hline
        \textbf{SlideSLAM} & 2024 & \boldcheckmark & \boldcheckmark & \boldcheckmark & \boldcheckmark & \boldcheckmark & \boldcheckmark \\
        \hline
        % Maggio et al. Clio \cite{tao20243d} & 2024 &  & &  &  & & \boldcheckmark & \boldcheckmark \\
        % \hline
        Tao et al. 3D Active \cite{tao20243d} & 2024 &  & & \boldcheckmark & \boldcheckmark & & \boldcheckmark \\
        \hline
        Chang et al. Hydra-Multi \cite{chang2023hydra-multi} & 2023 & \boldcheckmark & &  \boldcheckmark & \boldcheckmark & \boldcheckmark &  \\
        \hline
        Liu et al. Active Metric-Semantic \cite{liu2023active} & 2023 & \boldcheckmark & & \boldcheckmark & & \boldcheckmark & \boldcheckmark \\
        \hline
        Wu et al. An Object SLAM \cite{wu2023object} & 2023 & & & \boldcheckmark & \boldcheckmark &  \boldcheckmark & \\
        \hline
        Tian et al. Kimera-Multi \cite{Tian2022KimeraMulti} & 2022 & \boldcheckmark &  \boldcheckmark & & & & \\
        \hline
        Liu et al. Large-Scale \cite{liu2022large} & 2021 & & & \boldcheckmark & & \boldcheckmark & \boldcheckmark \\
        \hline
        Shan et al. Orc-VIO \cite{shan2020orcvio} & 2020 & & & \boldcheckmark & & \boldcheckmark & \\
        \hline
        Yang et al. Cube SLAM \cite{cubeslam_yang} & 2019 & & &  \boldcheckmark & &  \boldcheckmark & \\
        \hline
        Nicholson et al. Quadric SLAM \cite{Nicholson2019QuadricSLAM} & 2019 & & & \boldcheckmark & & \boldcheckmark& \\
        \hline
        Choudhary et al.   Distributed Mapping \cite{choudhary2017distributed} & 2017 & \boldcheckmark & \boldcheckmark & \boldcheckmark & & & \\
        \hline
       Bowman et al. Probabilistic \cite{bowman2017probabilistic} & 2017 & & & \boldcheckmark & & & \\
        \hline
        Salas-Moreno et al. SLAM++ \cite{SLAM++} & 2013 & & & \boldcheckmark&  & \boldcheckmark & \\
        \hline
        Cunningham et al. DDF-SAM \cite{cunningham2010ddf, cunningham2013ddf2} & 2013 & \boldcheckmark & \boldcheckmark & & & & \\
        \end{tabular}
        \arrayrulecolor{original}
        }
    \label{table:related-work-taxonomy-table}
\end{table*}
%%%%%%%%%%%%%%%%%%%%%%%%%%%%%%%%%%%%%%%%%%%%%%%%%%%%%%

{Motivated by this, recent advances in SLAM have pushed toward constructing metric-semantic maps \cite{bowman2017probabilistic, Nicholson2019QuadricSLAM, cubeslam_yang, rosinol2021kimera-new-ijrr, gu2024conceptgraphs}. However, they are typically limited to single-robot setups. For scalability and efficiency, an ideal system should support deployment across heterogeneous robot teams, enabling effective collaboration without relying on centralized communication or global localization infrastructure. Extending single-robot metric-semantic SLAM to  decentralized operation across multiple heterogeneous robots introduces several unique challenges. 
These include performing efficient place recognition and map merging among robots, maintaining consistent and probabilistically sound sensor measurement fusion under intermittent communication, and enabling collaboration among heterogeneous platforms. 
}

{To address these challenges, we} develop a multi-robot decentralized metric-semantic SLAM system that is open-sourced\footnote{Code: \url{https://github.com/KumarRobotics/SLIDE_SLAM}. Project website: \url{https://xurobotics.github.io/slideslam/}.} and suitable for real-time autonomous navigation and exploration {onboard resource-constrained single-robot as well as heterogeneous multi-robot platforms. We enable this by mapping the environment using a sparse map representation that explicitly models objects using simple shapes and can be used for task planning and long-horizon SLAM. 
The proposed sparse object-level map representation offers numerous advantages. First, it avoids the high runtime and memory demands of using raw point clouds or images, which is particularly important for large-scale missions with \gls{swap} constrained robot teams. }
Second, the lightweight nature of our representation ensures that robot-to-robot information sharing is feasible, even with limited communication bandwidth. This, coupled with our semantics-driven place recognition and loop closure algorithm, enables efficient detection of inter-robot loop closures.
Third, it provides a direct and intuitive representation for robots to accomplish high-level semantically meaningful tasks, such as actively exploring to search and reduce uncertainties in objects of interest. 
Fourth, unlike traditional approaches based on dense geometric features that usually have to marginalize variables frequently, which unavoidably leads to loss of information, our representation enables us to keep track of actionable information over a much larger scale. 
{Finally, it provides a unified representation across different sensing modalities. }

We summarize our \textbf{contributions} as follows.

\textit{Algorithm:} We develop a real-time decentralized metric-semantic SLAM framework that supports heterogeneous aerial and ground robots, which includes 
\begin{enumerate}[label=(\alph*)]
    \item {a computationally efficient back-end that uses our object-level metric-semantic map representation, and a flexible front-end that supports both LiDAR and RGBD sensors},
    \item semantics-driven place recognition algorithms that use sparse object-level maps for map merging,
    \item a decentralized multi-robot collaboration module that facilitates information sharing, even under intermittent communication conditions.
\end{enumerate}

\textit{System integration:} We integrate and deploy the proposed framework on a heterogeneous team of robots, demonstrating its capacity to enable semantics-in-the-loop autonomous navigation and exploration in various indoor and outdoor environments. The system operates in real time onboard SWaP-constrained robots, while maintaining moderate computation and memory demands.

\textit{Experiments:} We conduct extensive real-world experiments and provide thorough empirical results and analysis that highlight the efficiency, accuracy, and robustness of our system. We have also made our framework available to the public.

\section{Related Work}

\label{sec:related-work}

In this section, we categorize the related works into four key areas: metric-semantic SLAM, place recognition, multi-robot SLAM, and semantics-in-the-loop navigation. \cref{table:related-work-taxonomy-table} provides a quick summary of several key related works. The rest of this section provides a more comprehensive overview of the related works in each category.

%%%%%%%%%%%%%%%%%%%%%%%%%%%%%%%%%%%%%%%%%%%%%%%%%%%%%
\begin{figure*}[t!]
        \centering
        \includegraphics[trim=0 0 0 0, clip, width=1.0\textwidth]{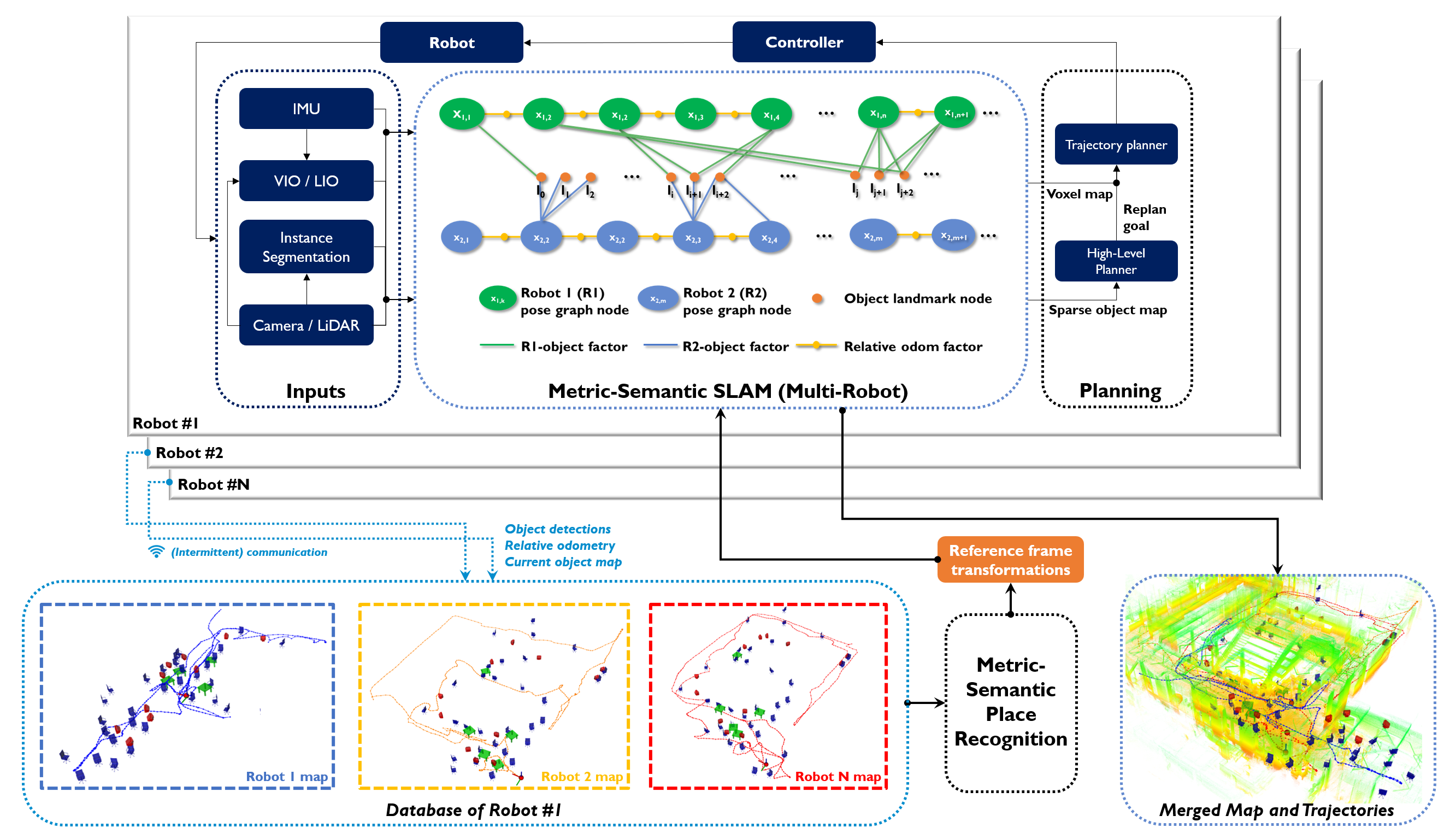}
        \caption{\textit{System Diagram.} Our system takes in data streams from each robot's onboard sensors, which can be either an RGBD camera or a LiDAR, and performs instance segmentation to extract semantic object features. Meanwhile, low-level odometry, either Visual-Inertial Odometry (VIO) or LiDAR-Inertial Odometry (LIO), provides relative-motion estimates between consecutive key poses. Next, the metric-semantic SLAM algorithm takes in such semantic observations and relative motion estimates, and constructs a factor graph consisting of both robot pose nodes and object landmark nodes. Meanwhile, our multi-robot communication module (see \cref{fig:decentralized-slam-collaboration-module}) opportunistically leverages connectivity to share lightweight semantic observations among robots in a decentralized way. Based on this shared information, our metric-semantic place recognition algorithm constantly checks for possible inter-robot loop closures at a fixed rate. Once a loop closure is detected, the resulting transformation between each pair of robots is used to transform all observations into each robot's reference frame. These observations are then added to their own factor graphs, forming a merged metric-semantic map. Note that the entire perception-action loop runs in a decentralized manner onboard each robot. Besides the obvious differences in control algorithms, the planning modules and the front-end processing algorithms are also different across each robot platform. This is due to the need to accommodate the differences in sensing modalities (RGBD and LiDAR), operating environments (indoor, urban, and forest), and traversal modes (ground and aerial). However, the core metric-semantic SLAM framework remains the same.}
        % \vspace{-0.1in}
        \label{fig:system-diagram}
\end{figure*}
%%%%%%%%%%%%%%%%%%%%%%%%%%%%%%%%%%%%%%%%%%%%%%%%%%%%%

\subsection{Metric-semantic SLAM}
Unlike traditional SLAM, metric-semantic SLAM constructs a map that encodes geometric features and semantic information on objects of interest. 
Metric-semantic SLAM has gained significant success and popularity in the past decade \cite{cadena2016past}, driven by the rapid development of deep learning techniques that extract semantic information from sensor data.
A variety of map representations are used in existing metric-semantic SLAM literature. 
Some use dense semantic maps, such as meshes \cite{rosinol2021kimera-new-ijrr}, volumetric maps \cite{grinvald2019volumetric, fusion++}, surfels \cite{chen2019suma++}, and 2.5D grid maps \cite{2.5DMapAutonomous, miller2021any, miller2022stronger}. 
Others use sparse object-level maps with prior information of object shapes, such as centroids \cite{bowman2017probabilistic}, cubes \cite{cubeslam_yang}, ellipsoids \cite{Nicholson2019QuadricSLAM}, cylinders \cite{sloam}, structured object models with prior shape constraints \cite{atanasov2018unifying, shan2020orcvio}, and mesh-based object models \cite{Feng2019MeshModel, SLAM++}. \todo{Following the recent trends in neural implicit representation, some prior works \cite{fu2023neuse, sucar2021imap, zhu2022nice} have also incorporated it into a map representation capable of metric-semantic SLAM, where objects are modeled implicitly using latent features from neural networks and used as localization constraints and for object association.} 
While dense metric-semantic maps are suitable for obstacle avoidance \todo{and facilitate more fine-grained modeling of objects}, their computational and memory demands, especially in SWaP-constrained platforms, can be significant. 
Thus, sparse object-level maps are desirable for real-time downstream tasks such as active information gathering, object manipulation and multi-robot collaboration. 
This is because semantic information is necessary for robots to perceive their environment and execute tasks in a semantically meaningful manner. Sparsity helps robots reduce resource demands on computation, storage, and communication. 
\revise{A recent work in this space is ConceptGraphs \cite{gu2024conceptgraphs}, which builds open-vocabulary, semantically rich 3D object maps. However, ConceptGraphs is not designed for real-time, onboard operations on resource-constrained robots. Its detection and 3D object mapping pipelines require extensive computational resources. 
By contrast, our work seeks to} develop a metric-semantic SLAM framework that utilizes sparse explicitly modeled objects and supports real-time decentralized operations within a heterogeneous robot team.

\subsection{Place recognition}
\label{subsec: place recognition} 
Place recognition and loop closure address the challenge of identifying whether a robot revisits a previously mapped location. This enables odometry drift correction in single-robot systems (intra-robot loop closure) and map merging in multi-robot systems (inter-robot loop closure). 
Prevalent approaches typically use consistency graphs. These methods operate on the assumption that pairwise distances between points or landmarks remain unchanged across two candidate maps \cite{mangelson2018pairwise-unweighted, yang2020teaser-unweighted, lusk2021clipper, leordeanu2005spectral-weighted}. The largest set of consistent associations is then identified and used to estimate the transformation. However, this assumption is invalid with the object-level map representations, where object detection and localization errors introduce additional inconsistency. 
\todo{Some geometric-based place recognition methods, such as \cite{xu2023ring++}, design hand-crafted features for LiDAR point clouds that are translation and rotation invariant to match them accurately and establish robust place recognition. Other methods like \cite{cattaneo2022lcdnet} generate similar global matching features and perform relative pose refinement using deep neural networks. While these methods demonstrate impressive place recognition performance, they may have high memory usage when storing a large number of geometric features or require large amounts of data to fine-tune the neural network for place recognition in novel environments.}
Conversely, semantic maps are much sparser, provide richer information and can be abstracted from any sensing modality, LiDAR or camera. Some approaches leverage this fact and propose descriptor-based methods based on the spatial relationship of semantic objects. 
In \cite{XViewGS}, a random walk descriptor encoding semantic labels of neighboring nodes is designed for each semantic node. 
A semantic histogram-descriptor-based method is proposed in \cite{histogramGraphMAtch}, which improves the efficiency of graph matching compared to the random-walk-based method. 
\cite{wu2023object} also leverages random walk descriptors with semantic information but further adds information on object parameters, angles, and distance between an object node and its neighboring nodes to filter out false candidates. 
In \cite{nardari2020place}, Urquhart tessellations are derived based on the positions of semantic landmarks (tree trunks in forests). Polygon-based descriptors are then computed to describe the local neighborhood of the robot, which are then used for place recognition. 
Hierarchical descriptors consisting of appearance-based descriptors like Distributed Bag of Words (DBoW2) \cite{bow2} and higher level descriptors encoding information about a node's neighboring objects or places are designed in \cite{hughes2022hydra, chang2023hydra-multi}. 
Drawing inspiration from some of these prior works, we design two complementary place recognition algorithms that leverage our metric-semantic map representation \todo{from diverse sensing modalities (LiDAR and RGBD Camera)} and adopt exhaustive search-based and descriptor-based strategies, respectively.

%%%%%%%%%%%%%%%%%%%%%%%%%%%%%%%%%%%%%%%%%%%%%%%%%%%%%
\begin{figure*}[t!]
    \centering
    \includegraphics[trim=0 0 0 0, clip, width=2.0\columnwidth]{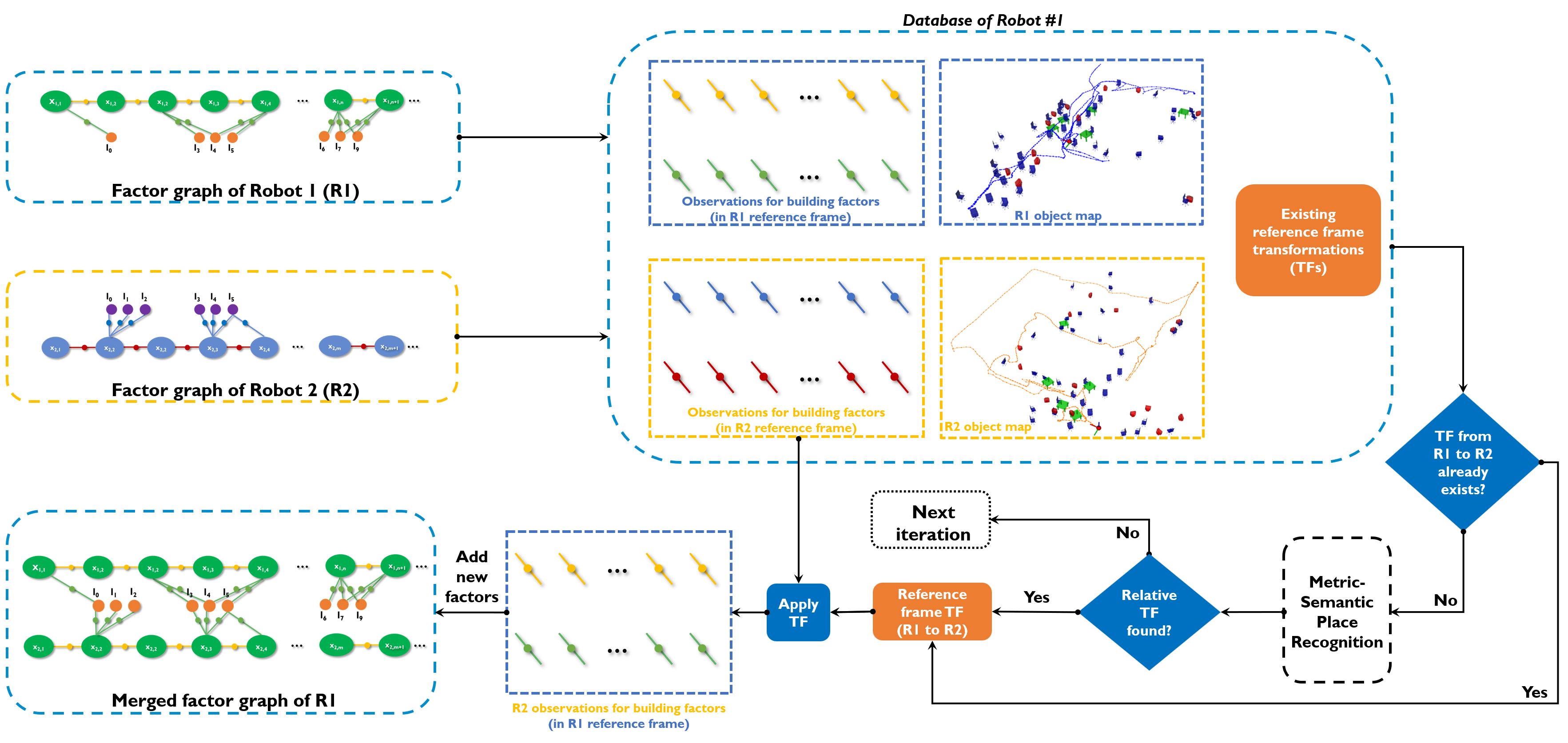}    
    \caption{\textit{Multi-robot collaboration module for decentralized metric-semantic SLAM.} The robots share lightweight metric-semantic observations necessary for constructing the factors between object landmarks and robot poses in the factor graph, which include the detected objects and the odometry relative motion estimate (w.r.t. the previous pose) associated with each key pose in the factor graph. Once the metric-semantic place recognition module successfully finds a loop closure with another robot, the shared observations from that robot will be transformed into the current robot's reference frame and added to the factor graph of the current robot.}
    \vspace{-0.1in}
    \label{fig:decentralized-slam-collaboration-module}
\end{figure*}
%%%%%%%%%%%%%%%%%%%%%%%%%%%%%%%%%%%%%%%%%%%%%%%%%%%%%

\subsection{Multi-robot SLAM}
Multi-robot SLAM systems expand upon their single-robot counterparts by incorporating two crucial modules: inter-robot loop closure and multi-agent graph optimization. 
Numerous studies address the challenge of place recognition, as detailed in \cref{subsec: place recognition}. 
Representative works in multi-robot SLAM include \cite{onlineMultiRobotSLAM, lampSLAM2}, which rely on a central base station or agent to merge measurements from multiple robots, and \cite{doorSLAM, discoSLAM, swarm-slam, data-efficient-slam}, which operate in a distributed or decentralized manner. 
Distributed multi-robot systems involve robots that coordinate through message passing, possibly with central nodes, while decentralized systems operate without a central decision point, enhancing robustness by allowing each robot to make independent decisions. 
Decentralized multi-robot SLAM is a promising framework in multi-robot SLAM research. 
It enables robots to operate independently and collaborate opportunistically when communication links are established, allowing them to accomplish complex tasks such as large-scale collaborative exploration without relying on infrastructure. 

However, the prior works mentioned above only contain geometric information during mapping. Recent efforts extend the field to metric-semantic mapping. \cite{Tian2022KimeraMulti} presents a system producing a dense 3D semantic mesh, incorporating distributed loop closure detection and distributed robust pose-graph optimization based on Riemannian block coordinate descent \cite{RBCD}. However, semantic information is used primarily in the map representation rather than for localization and loop closure. Another related work, \cite{choudhary2017multi}, proposes a distributed Pose Graph Optimization (PGO) framework based on a distributed Gaussian-Seidel approach \cite{DGS} by considering only overlapping constraints among robots in the optimization and extracting objects as landmarks. 
Object-level semantic landmarks, because they are viewpoint invariant, can facilitate collaboration among heterogeneous robots.
Few works exist to solve the problem of collaborative SLAM using heterogeneous sensors. Among them, \cite{posegraphSlamIsaac} fuses local pose graphs from different SLAM algorithms into a global pose graph. However, it is limited to vision-based sensors, which require sharing images associated with each pose for map merging. 
Our work proposes a more generic and efficient system towards decentralized metric-semantic SLAM that supports real-time operation on \gls{swap}-constrained heterogeneous robot platforms with either RGBD or LiDAR sensors.

\subsection{Semantics-in-the-loop navigation and exploration}
Most autonomous navigation systems rely on geometric maps. 
Although such maps are reliable for obstacle avoidance, they lack semantic information that robots can leverage to improve their state estimation and planning while navigating the environment. 
Object-based landmarks help minimize odometry drift, thus benefiting autonomous navigation if executed onboard in real-time \cite{liu2022large}. 
Recent work goes beyond improving odometry but uses semantic information for active uncertainty reduction planning \cite{georgakis2021learning, zheng2019active, liu2023active}, and multi-robot collaborative planning \cite{miller2022stronger}. 
The use of semantic information in active SLAM is relatively uncommon \cite{placed2023survey}. 
Some representative work in this field includes \cite{asgharivaskasi2021active, asgharivaskasi2023semantic}. 
However, there has not yet been a real-time decentralized metric-semantic SLAM framework that enables a team of heterogeneous robots to explore GPS-denied environments while reasoning about the semantic information and leveraging it to facilitate multi-robot collaboration. Our work seeks to develop such a framework.

\section{Problem Formulation}
\label{sec:problem-statement}

Given an unknown environment and a team of heterogeneous robots, our objective is to construct a hierarchical metric-semantic map of the environment without relying on any infrastructure. The robots can communicate with each other opportunistically. Each robot must estimate its pose and the state of the environment and use this information to determine actions for navigation and exploration in real time.

\subsection{Preliminaries}
Consider $K$ robots, indexed from 1 through $K$. The map of the $k$-th robot $\mathcal{M}^{(k)}$ consists of a set of object landmarks $\{\mathbf{\ell}^k_1, \ell^k_2, \ldots, \ell^k_{n_k}\}$ belonging to a set of pre-defined classes (i.e., object categories). 
An object is described by its class and a state vector, including its position, orientation, and shape information. 
For example, a cuboid has three shape parameters describing its width, length, and height. 
A cylinder has one parameter describing its radius (we do not estimate its height as it is unobservable for the majority of measurements due to limited field of view).
Ellipsoid models are simplified to have two shape parameters (radius and height) and known identity orientation. Objects of each class are modeled using one of the aforementioned three shapes as specified by the user. For each key pose of robot $k$, denoted by $\mathbf{x}^k_t$, the associated measurements include object detections from this pose and the relative motion from the previous key pose. The details of the shape models, robot measurements, and the factor graph used for optimization are presented in \cref{sec:ms-slam-framework-factors-measurements-models}.

\subsection{Dec-Metric-Semantic SLAM for multi-robot exploration}
\label{sec:Decentralized Metric-Semantic SLAM formulation}

In this section, we state the general active decentralized metric-semantic SLAM problem (Dec-Metric-Semantic SLAM).
Let $\mathcal{I}_{t}^{(k)}$ be all past observations that robot $k$ has accrued by time $t$, from its own measurements, as well as from messages it has received from other robots. 
At time $t$, every robot $k$ executes an exploration or navigation policy: 
\begin{equation}
    \mu_{t}^{(k)} : \mathcal{I}_{t}^{(k)} \rightarrow \mathcal{A}, 
\end{equation}
where $\mathcal{A}$ is an action set of robot $k$ that can include, for example, waypoints to follow or low-level control inputs. We formulate the exploration problem as:
\begin{equation}\label{eqn:problem1}
\max_{ ( \mu_{t}^{(k)} )_{\substack{1 \leq t \leq T,\\ 1 \leq k \leq K}} } \mathbb{E} \left[ \sum_{k = 1}^{K} \sum_{t = 1}^{T} \mathcal{U}_{t}(\mathbf{x}_{t}^{(k)}, \mathcal{M} ; \mathcal{I}_{t}^{(k)})\right],
\end{equation}
where $\mathcal{M}$ represents the (static) map of the environment, and $\mathbf{x}_{t}^{(k)}$ the state of robot $k$ at time $t$.
The function $\mathcal{U}_t$ is a (potentially) time-varying utility function that quantifies the accuracy with which a robot can estimate specified quantities of interest given all the information it has amassed thus far. 

In our metric-semantic SLAM formulation, the quantities of interest include the robot poses $\mathbf{x}_{t}^{(k)}$ and the map of the environment $\mathcal{M}$. At each time $t$, robot $k$ receives odometry measurements 
\begin{equation}
    \mathbf{h}_{t}^{(k)} = \mathbf{x}_{t}^{(k)} \ominus \mathbf{x}_{t-1}^{(k)} + \mathbf{w}_{t}^{(k)},
\end{equation} 
where $\ominus$ computes the relative pose between two key poses, and map measurements 
\begin{equation}
    \mathbf{m}_{t}^{(k)} = h(\mathbf{x}_{t}^{(k)}, \mathcal{M}) + \mathbf{v}_{t}^{(k)}.
\end{equation}
We assume $(\mathbf{w}_{t}^{(k)})_{t \leq T, k \leq K}$ and $(\mathbf{v}_{t}^{(k)})_{t \leq T, k \leq K}$ are independent random variables with zero mean and covariance $\Lambda_{k}$ and $\Gamma_{k}$, respectively. \todo{Currently, both variables $\mathbf{w}_{t}^{(k)}$ and $\mathbf{v}_{t}^{(k)}$ are set to constant values since key poses $\mathbf{x}_{t}^{(k)}$ are sampled at a fixed robot travel distance interval. However, in our implementation, we provide an interface to allow scaling $\mathbf{w}_{t}^{(k)}$ according to travel distance and modify $\mathbf{v}_{t}^{(k)}$ accordingly to account for sensor and object detection noise.}

We let $C_{t} \in \{0,1\}^{K \times K}$ be the time-varying symmetric adjacency matrix of the undirected graph modeling the robot's communication. At every time $t$, any distinct pair of robots $i$ and $j$ communicate if and only if $(C_{t})_{i,j} = (C_{t})_{j,i} = 1$. We denote the message that robot $i$ sends to robot $j$ at time $t$ via $r_{(i \rightarrow j)}(t)$. We denote the neighbors of any robot $k$ in the graph at time $t$ via 
\begin{equation}
\mathcal{N}_{t}^{(k)} = \{ j \in [K] \setminus \{k\} \ : \ (C_{t})_{j,k} = 1\},
\end{equation}
where $[K]:=\{1,\ldots,K\}$. For any distinct pair of robots $(i,j)$, we denote their last meeting time up to time $t$ as 
\begin{equation}
L_{t}^{(i,j)} = \max \{ s \leq t \ : \ (C_{s})_{i,j}=1 \},    
\end{equation} 
noting the symmetry in the superscript, i.e., $L_{t}^{(i,j)} = L_{t}^{(j,i)}$. 
Then, the information state of robot $k$ may be defined recursively as 
\begin{equation}
    \label{eqn: information merge union}
    \mathcal{I}_{t}^{(k)} = \bigcup_{j \in [K] \setminus \{k\}} \mathcal{I}_{L_{t}^{(k,j)}} \ \cup \ \bigcup_{s \in [t]} \left(\{\mathbf{h}_{s}^{(k)} \} \ \cup \ \{\mathbf{m}_{s}^{(k)}\}\right),
\end{equation}
with initial condition $\mathcal{I}_{0}^{(k)} = \emptyset$ for all $k \in [K]$.
%%%%%%%%%%%%%%%%%%%%%%%%%%%%%%%%%%%%%%%%%%%%%%%%%%%%%%%%%%%%%%%%%

We develop an approach that uses a set of policies $\mu$ (e.g., active SLAM, exploration, etc.) that take as input compressed representations of $\mathcal{M}$ using a custom data structure, i.e., a set of object landmarks with corresponding poses, shapes, and class labels, as well as a custom rule for updating the data structure both using new measurements from the robot and information from its communication with other robots. The exploration or navigation policy of robot $k$ may then be defined via
\begin{equation}
\mu_{t}^{(k)}(\mathcal{I}_{t}^{(k)}) := \mu_{t}^{(k)}( \Pi (\mathcal{I}_{t}^{(k)}) ),
\end{equation}
where the information projection operator $\Pi$ is given by
\begin{equation}
\Pi( \mathcal{I}_{t}^{(k)} ) = ( \hat{\mathbf{x}}_{1:t}^{(k)}, \hat{\mathcal{M}}_{t}^{(k)} ).
\end{equation} 
$\Pi$ extracts an estimate of the previous trajectory of robot $k$, as well as its estimate of the map via 
\begin{equation}
\hat{\mathbf{x}}_{1:t}^{(k)}, \hat{\mathcal{M}}_{t}^{(k)} = \underset{\mathbf{x}_{1:t}^{(k)}, \mathcal{M}}{\text{argmax}} \ P( \mathbf{x}_{1:t}^{(k)}, \mathcal{M} \ \vert \ \mathcal{I}_{t}^{(k)} ),
\label{eqn:problem2-mle}
\end{equation}
where $P( \mathbf{x}_{1:t}^{(k)}, \mathcal{M} \ \vert \ \mathcal{I}_{t}^{(k)} )$ represents the joint probability distribution of the robot $k$'s trajectory from the start up to time $t$, and the map $\mathcal{M}$, which includes object landmarks, conditioned on the information state $\mathcal{I}_{t}^{(k)}$ of the robot $k$. 
Finally, we use the following form of messages
\begin{equation}
r_{(i \rightarrow j)}(t) = (\mathcal{I}_{t}^{(i)}, \hat{\mathcal{M}}_{t}^{(i)}).
\end{equation}

In principle, robot $j$ can construct $\hat{\mathcal{M}}_{t}^{(i)}$ from $\mathcal{I}_{t}^{(i)}$. Given the fact that the compressed representation $\hat{\mathcal{M}}_{t}^{(i)}$ we use is memory efficient, we directly share this information to avoid the extra computation brought about by the reconstruction step. The map $\hat{\mathcal{M}}_{t}^{(i)}$ is shared only for checking inter-robot loop closures and estimating the transformations. Once a valid inter-robot loop closure is established, $\mathcal{I}_{t}^{(i)}$ is used to form the updated $\mathcal{I}_{t}^{(j)}$ as mentioned in \cref{eqn: information merge union}. $\mathcal{I}_{t}^{(j)}$ will then be used to update $\hat{\mathcal{M}}_{t}^{(j)}$.

\section{Metric-Semantic SLAM}
\label{sec:ms-slam-framework-factors-measurements-models}

\subsection{Approach overview}
\label{sec:approach overview}

We next turn to an overview of the approach for computing the information projection operator $\Pi$, which is one of the core contributions of this paper. 
Unlike traditional geometric-only methods such as point cloud registration, our method leverages both \textit{geometric} and \textit{semantic} information to find common components of maps estimated by different robots, and then refines such estimates using both its own measurements and measurements made by other robots.
We illustrate this idea with a pair of robots $k$ and $j$ at time $t$, though the argument can be readily generalized to simultaneous interactions of a larger number of robots. 
Robot $k$ solves the maximum likelihood state estimation problem at time $t$ as follows. For every robot $j \in \mathcal{N}_{t}^{(k)}$, we consider the message $r_{j \rightarrow k}(t) = (\mathcal{I}_{t}^{(j)}, \hat{\mathcal{M}}_{t}^{(j)})$. 
First we determine common map observations made by robots $k$ and $j$ up to time $t$ by performing inter-robot loop-closure detection on $\hat{\mathcal{M}}_{t}^{(j)}$ and $\hat{\mathcal{M}}_{t}^{(k)}$. 
The output of the loop-closure module yields an estimate of the relative transformation ${}^{(k)}T_{(j)} = ({}^{(k)}R_{(j)}, {}^{(k)}\mathbf{t}_{(j)}) \in \mathbb{SE}(3)$ between reference frames of robot $j$ and robot $k$. 
In particular, a point with coordinates ${}^{(j)}\textbf{p}$ in the reference frame of robot $j$ has coordinates ${}^{(k)}\textbf{p} = {}^{(k)}R_{(j)} {}^{(j)}\textbf{p} + {}^{(k)}\mathbf{t}_{(j)}$ in the reference frame of robot $k$.
This transformation is then used to determine which landmarks in the map of robot $j$ correspond to landmarks that also exist in the map of robot $k$. 
Thereafter, the trajectories of both robot $j$ and $k$ are optimized together with the union of object models in the map, taking care to associate measurements of the same landmark taken by two robots with the same variable in the factor graph:
\begin{equation}
\begin{aligned}
\hat{\mathbf{x}}_{1:t}^{(j)}, \hat{\mathbf{x}}_{1:t}^{(k)}, \hat{\mathcal{M}}^{(k)}_t = 
\underset{\mathbf{x}_{1:t}^{(j)}, \mathbf{x}_{1:t}^{(k)}, \mathcal{M}}{\text{argmax}} \ P( \mathbf{x}_{1:t}^{(j)}, \mathbf{x}_{1:t}^{(k)}, \mathcal{M} \ \vert \ \mathcal{I}_{t}^{(k)} ).
\end{aligned}
\end{equation}
The above equation provides details on the information projection operator $\Pi$.

Our system works on robots with different types of sensors, including RGBD cameras and LIDARs, as illustrated in \cref{fig:system-diagram} and detailed in \cref{sec:system overview}. Our metric-semantic SLAM framework handles such heterogeneity by using different front-end pipelines. We integrate the proposed metric-semantic SLAM framework with our autonomous exploration and navigation stack as detailed in \cref{sec:integration-with-autonomy}. In the rest of this section, we will provide details on individual modules of the proposed framework.

\subsection{Map representation} 
\label{subsec: desired attributes and map representation of the framework}

\begin{table}[h!]
\centering

        \caption{\textit{Requirements for map representation.} Generic implies that the map must be adaptable to various sensing modalities and environments. Informative means that the map should contain both metric and semantic information. Sparse indicates that the map needs to be memory efficient for storage and sharing.}
\resizebox{0.5\textwidth}{!}{
\begin{tabular}{||c|c|c|c||}
\hline
& \begin{tabular}[c]{@{}c@{}}Heterogeneous robots \\ and environments\end{tabular} & \begin{tabular}[c]{@{}c@{}}Place recognition \\ and loop closure\end{tabular} & \begin{tabular}[c]{@{}c@{}}Real-time explora- \\ tion and navigation\end{tabular} \\ \hline
\hline
Generic     & \boldcheckmark         & \boldcheckmark      &                                      \\ \hline
Informative &                                       & \boldcheckmark      & \boldcheckmark        \\ \hline
Sparse      & \boldcheckmark         & \boldcheckmark      & \boldcheckmark        \\ \hline
\end{tabular}
}
        \label{table:Requirements for map representation.}
\end{table}

The metric-semantic SLAM framework needs to satisfy several design attributes. These include (1) support for heterogeneous robot teams, where the robots may carry different sensors, operate in different environments, and share information with each other via intermittent communication, 
(2) support for inter-robot place recognition for map merging,
(3) capability of real-time operation under the constraints of onboard computation and memory resources, so that the robot can use the estimated pose and the metric-semantic map to guide its navigation. 

As summarized in \cref{table:Requirements for map representation.}, to support different kinds of sensors (RGBD cameras and LiDARs) and objects in different environments, we require the map representation to be generic. To work with multiple robots and allow the sharing of information with limited communication bandwidth, we require the map representation to be sparse. To enable efficient and accurate place recognition and loop closure, we additionally require the map representation to be informative so that the robots can distinguish different semantic objects even under perceptually aliased conditions. Finally, to enable efficient, large-scale, and real-time autonomous exploration, we also require the map to be sparse and informative. The informativeness allows the exploration planner to better understand the environment, such as the uncertainties in the semantic objects, and generate informative paths.

Therefore, we design a generic, sparse, and informative metric-semantic map representation. It contains a set of objects, each represented by a semantic class and a state vector that describes its model parameters, including position, orientation, and shape information. Such a map representation, as illustrated in \cref{fig:seven-robot-indoor-outdoor}, can be maintained throughout the entire mission of the robot \todo{while enabling tasks such as localization, mapping, map sharing, place recognition, map merging, and active exploration over a large scale and in real time.}

% \begin{figure}[b!]
%     \centering
%     \includegraphics[trim=0 0 0 200, clip, width=0.45\columnwidth]{img/final-high-res-figures/new_autonomu_in_the_loop.png}
%     \caption{\textit{Semantics-in-the-loop autonomy.} The Falcon 250 UAV autonomously explores the environment while constructing a voxel map utilized for low-level navigation and a metric-semantic map used for active semantic loop closure in real time. The long-horizon exploration trajectory (thin green line) is shown alongside the dynamically feasible trajectory (thick blue curve) planned toward the next exploration waypoint.}
%     \label{fig:Visualization of semantic-in-the-loop autonomy}
% \end{figure}

The metric-semantic SLAM problem can be broken down into two subproblems: (a) determining the discrete semantic labels and data association of detected objects and (b) optimizing over the continuous variables of robot poses and object model parameters (pose and shape). In this work, we approach problem (a) using deep neural networks for detection and assignment algorithms for data association, and problem (b) by first converting object observations into factors in a factor graph and then using an incremental smoothing and mapping algorithm (iSAM2) \cite{kaess2012isam2, gtsam} to optimize it. In the rest of this section, we provide details on these two steps. 

\subsection{Object detection and modeling}

The front-end of our SLAM framework is responsible for processing raw data from different sensors and converting it into object-level observations. 
The process is broken down into three components: 
1) object detection or instance segmentation; 
2) object instance tracking to accumulate observations from different views;
3) shape model fitting to the instances based on the class labels.

% \subsubsection{Closed-set object detection}
For semantic segmentation on point cloud from LiDAR, we trained RangeNet++ \cite{milioto2019rangenet++} with a small modified backbone. This model performs segmentation on the range image, which is a spherical projection of a point cloud. 
This operation drastically decreases the inference time compared to performing inference on a 3D data structure such as a point cloud and makes real-time inference onboard \gls{swap}-constrained platforms possible. 
We use a custom open-source labeling tool to generate training data  \cite{faster-sill}. 

For the RGBD sensor, we use YOLOv8 \cite{Jocher_YOLOv8_by_Ultralytics_2023} to perform instance segmentation on RGB images and backproject the pixels in each object instance's segmented mask into point clouds using depth information. 
We apply depth-based thresholding within each object instance's mask, keeping only the points within a certain range of depth percentile, to avoid including noisy pixels.

\revise{ 
In addition to supporting the aforementioned closed-set object detection methods that only enable the robot to perceive a subset of the semantic information in the world characterized by pre-identified labels, 
we provide support to incorporate open-vocabulary object detectors into our system. 
Specifically, we utilize the YOLO-World model \cite{cheng2024yolo-world} as our open-vocabulary object detector due to its real-time onboard performance.} The YOLO-World model processes pairs of images and a list of text prompts, where we specify a set of object categories of interest. The model then outputs object detections corresponding to each query in the text prompts. We apply depth thresholding to these detections, similar to the approach used for YOLOv8.

Once the point cloud per detection is extracted, these detected objects are tracked over time using the Hungarian assignment algorithm \cite{kuhn1955hungarian}. 
Tracking instances across time and from different viewpoints ensures robust geometry of objects and minimizes false positives by rejecting those tracks that only appeared once or very few times, which is necessary for metric-semantic mapping. 
Finally, the algorithm also helps with robustness against moving objects by tracking and accumulating their point clouds. 
For example, an accumulated instance of a moving object appears elongated when compared to its true size, and such objects can then be filtered out by simply applying a threshold on their dimensions. 
This preliminary tracking step is used to more robustly obtain object-level observations.
In the factor graph optimization back end, there is a separate object tracking step that associates these object observations with existing object landmarks to construct factors in the factor graph.

Once a robust object instance is obtained, depending on the type of object, a shape model is fitted to the instance. Objects such as vehicles use a cuboid model, tree trunks and light poles use a cylinder model, and irregularly shaped objects that cannot be properly categorized into either a cuboid or cylinder use an ellipsoid model. Having determined an appropriate model for the object instance, these models are converted into factors to be added to the factor graph for the SLAM back-end optimization.

\revise{We highlight that we do not rely on a specific object detector. Our implementation has a modular front-end design, enabling the seamless integration of alternative object detectors in the future.}

\subsection{Factor graph optimization with object models}
\label{subsec:MS-SLAM Backend}

After acquiring the object detections associated with each pose of the robot, we now formulate the back-end optimization problem for the metric-semantic SLAM. 
When a new object detection arrives, we first check if we can associate it with existing object landmarks in the map. 
Upon the first observation of the object, we will initialize an object landmark. After that, once we associate new object models with existing object landmarks, we use the detection-landmark matches to form factors between the current pose and the matched landmarks in the factor graph.  
A valid object detection-landmark association needs to meet two criteria: (1) they have the same semantic label, and (2) their distance and model difference are within a certain threshold. 

A critical step to form such factors is to define the constraining relationship (i.e., graph edges) between the object and robot pose graph vertices. 
This means we need to define the measurement models that convert observations of relative odometry and object models into factors. We denote the matrix form of the robot pose $\mathbf{x}_t$ as $\mathbf{H}^w_s$, where ${\mathbf{R}^w_s}$ is the rotational component and $\mathbf{t}^w_s$ is the translational component. In the rest of this section, we will describe this process in detail. 

\subsubsection{Cuboid factors}

First, we define the state vector of the cuboid object model as: $\ell^{g} = [\mathbf{r}; \mathbf{t}; \mathbf{d}]$, where $\mathbf{r}$ = [$r_x$,  $r_y$,  $r_z$]$^\intercal$ is the rotation vector, 
$\mathbf{t}$ = [$t_x$,  $t_y$,  $t_z$]$^\intercal$ is the translation vector, and 
$\mathbf{d}$ = [$d_x$,  $d_y$,  $d_z$]$^\intercal$ is the size.

 We first recover the cuboid's $\mathbb{SE}$(3) pose based on its state vector $\ell_i^{g} = [\mathbf{r}; \mathbf{t}; \mathbf{d}]$, and transform it from the reference frame to the body frame using ${\mathbf{H}^s_\texttt{cub}} = \mathbf{H}^s_w \mathbf{H}^{w}_\texttt{cub}$. The error function for cuboidal objects is as follows:
\begin{equation}
\mathbf{e}_\texttt{cub} = 
\begin{bmatrix}
\mathbf{log}((\mathbf{H}^{s}_\texttt{cub}(\mathbf{z}))^{-1} ({\mathbf{H}^s_w} \mathbf{H}^{w}_\texttt{cub}))^{\vee} \\ \mathbf{d} - \mathbf{d}{(\mathbf{z})}
\end{bmatrix},
\end{equation}
where $\vee$ is vee operator that maps the $\mathbb{SE}$(3) transformation matrix into  $6\times1$ vector, $\mathbf{log}$ is the log map, and $(\cdot){(\mathbf{z})}$ are corresponding measurements.

\subsubsection{Cylinder factors}

First, we define the state vector of the cylinder object model as: 
$\ell_i^{g} = [\mathbf{b}; \mathbf{n}; r]$,
where $\mathbf{b}$ = [$b_x$,  $b_y$,  $b_z$]$^\intercal$ is the origin of the axis ray,
$\mathbf{n}$ = [$n_x$,  $n_y$,  $n_z$]$^\intercal$ is the direction of the axis ray, and 
$r$ is the radius. 

Similarly, the treatment of cylinder factors has been derived in \cite{liu2023active}. We calculate the expected measurement and actual measurement of cylinder objects from $\ell_i^{g}$ and $\ell_i^{g} (\mathbf{z})$. We define the error function for cylindrical objects as:
\begin{equation}
\mathbf{e}_\texttt{cyl} 
= 
\begin{bmatrix}
(\mathbf{R}^s_w\mathbf{b} + \mathbf{t}^s_w) - \mathbf{b}{(\mathbf{z})}\\ 
\mathbf{R}^s_w\mathbf{n} - \mathbf{n}{(\mathbf{z})}\\ 
{r} - {r}{(\mathbf{z})}
\end{bmatrix},
\end{equation}
where $(\cdot){(\mathbf{z})}$ are corresponding measurements.

\subsubsection{Ellipsoid factors}
All objects that are not associated with a cylinder or cuboid shape are represented by ellipsoid landmarks. 
Note that our ellipsoid model is simplified to have only two dimensions (i.e., the cross-section is a circle) and identity orientation.  For each ellipsoid object,  the state vector is: $\ell_i^{g} = [\mathbf{c}, \mathbf{d}_e]^\intercal$, where $\mathbf{c}$ = [$c_x$,  $c_y$,  $c_z$]$^\intercal$ represents 3D position of the centroid, and $\mathbf{d}_e$ = [$d_r$,$d_h$]$^\intercal$ represents the radius and height.

Given the expected ellipsoid landmark model $\ell_i^{g} = [\mathbf{c}, \mathbf{d}_e]^\intercal$ and the actual measurements which consists of range-bearing measurements $rg{(\mathbf{z})}, \theta{(\mathbf{z})}, \phi{(\mathbf{z})}$ as well as the dimension measurements $\mathbf{d}_e(\mathbf{z})$ = [$d_r(\mathbf{z})$,  $d_h(\mathbf{z})$]$^\intercal$, the measurement error $ \mathbf{e}_{\texttt{ellip}}$ is derived as follows:

Let $[c_x',  c_y',  c_z'] = \mathbf{R}_t^\top (\mathbf{c} - \mathbf{t}_t)$ denote the landmark transformed into the body frame, $rg_{\text{exp}} = \sqrt{{c_x'}^2 + {c_y'}^2 + {c_z'}^2}$ represent the expected range measurement, and $\theta_{\text{exp}} = \tan^{-1}\left(\frac{c_y'}{c_x'}\right)$ and $\phi_{\text{exp}} = \tan^{-1}\left(\frac{c_z'}{\sqrt{{c_x'}^2 + {c_y'}^2}}\right)$ indicate the expected bearing measurements. The measurement error for centroids of ellipsoid objects is as follows:
\begin{equation}
\mathbf{e}_{\texttt{ellip}} = \begin{bmatrix}
rg_{\text{exp}} - rg{(\mathbf{z})} \\
\theta_{\text{exp}} - \theta{(\mathbf{z})} \\
\phi_{\text{exp}} - \phi{(\mathbf{z})} 
\end{bmatrix}.
\end{equation}
Since the GTSAM solver~\cite{gtsam} already provides range and bearing measurement models, we directly utilize these for optimizing the centroid positions. For dimension estimation, we use the moving average method as follows: 
$\mathbf{d}_e^{updated} = (1 - \alpha) \cdot \mathbf{d}_e + \alpha \cdot \mathbf{d}_e(\mathbf{z})$, where $\alpha$ controls the weight given to new measurements.

\subsubsection{Odometry factors}

The robot state at time $t$ is represented by $\mathbf{x}_t$, which contains the $\mathbb{SE}$(3) pose.  For each key pose $\mathbf{x}_t$, we compute the relative motion from its preceding key pose using the low-level odometry readings from either VIO or LIO, depending on the sensor that the robot carries. Specifically, this is calculated as $\Delta \mathbf{x}_t = (\mathbf{x}_{t-1}^{\text{odom}})^{-1} \circ \mathbf{x}_t^{\text{odom}}$. Our factor graph incorporates this relative motion as an odometry factor. This approach mitigates the cumulative drift commonly associated with low-level odometry systems by not directly using the low-level odometry as pose priors. Instead, we utilize the odometry estimates to interpolate between consecutive key poses that are proximate in both space and time, where the precision of the low-level odometry provides reliable measurements.

We use the measurement models defined above to convert the observations and build a factor graph with nodes for both robot poses and object landmarks and factors between them, as illustrated in \cref{fig:system-diagram}. 
We implemented our custom cuboid and cylinder factors to be compatible with the GTSAM factor graph optimization library~\cite{gtsam}. In addition, we use GTSAM's built-in relative $\mathbb{SE}$(3) pose measurement model for the odometry factor and the range-bearing measurement model for ellipsoid object landmarks.

\subsection{Place recognition and loop closure}

Our system is designed with a modular architecture, enabling users to integrate alternative place recognition and loop closure algorithms to meet their unique requirements. 
In this paper, we have developed two example algorithms that leverage our object-level metric-semantic map representation. At a high level, they can be summarized as follows: 
\begin{enumerate}
    \item \textit{SlideMatch}: This approach uses an exhaustive search strategy by sampling candidate transformations, applying them to the first metric-semantic map, and matching the transformed map against the second map to identify potential loop closures. 
    \item \textit{SlideGraph}: This descriptor-based approach begins by performing data association through the matching of triangle descriptors derived from each metric-semantic map, using both class labels and position information of the objects. The initial associations are then refined using a graph-theoretic framework \cite{lusk2021clipper}, which effectively eliminates outliers. 
\end{enumerate}
The two example algorithms exhibit different behaviors and offer complementary performance across various scenarios. Generally, we recommend using SlideGraph as the primary algorithm, with SlideMatch serving as an alternative option, as further discussed in detail in \cref{subsec:slidematch vs. slidegraph}.

\subsubsection{SlideMatch}
\label{subsec: slidematch}

The SlideMatch algorithm takes a pair of object-level metric-semantic maps from either the same robot (i.e., historical map and current map for loop closure) or a pair of different robots.  SlideMatch algorithm checks within a search region if a valid loop closure is found and outputs the relative transformation between the reference frames of these two maps. 
Unlike \cite{Prabhu2024-UAVs-for-forestry}, here we also leverage additional information such as semantic labels and dimensions of objects in the maps. 
The search region of the algorithm encompasses continuous ranges of X and Y positions and yaw angles, where the relative transformation between the two maps may exist. The search procedure is carried out over the search region using a user-defined search resolution for discretization.

\textit{Preprocessing:} For inter-robot place recognition, 
we designed a preprocessing step that zero-centers the two object-level maps before performing data association. 
This step reduces the search region and computation of the SlideMatch algorithm, especially when the two robots start far apart. Furthermore, it enables the algorithm to automatically calculate the search region by setting the X and Y search ranges to encompass the region where the two maps overlap and the yaw search range to $(-\pi,\pi]$. 

\textit{Anytime implementation:} The anytime implementation progressively improves the quality of place recognition until the user-specified compute budget is exhausted. 
 Quality is measured by the number of inlier matches. 
This approach guarantees the algorithm returns the best estimate within the given computation time budget. A valid loop closure is considered to be found if the matching score of the best estimate exceeds the valid loop closure inlier threshold, which \todo{is described in more detail below}. This anytime procedure is achieved by gradually increasing the search region for data association until it covers the entire region or the compute budget is used up.

\textit{Data association:} The data association step involves finding the possible set of landmarks that are common in both maps. Candidate transformation samples are generated from the search region by discretizing the search space across the X, Y, and yaw dimensions, based on the user-specified resolutions. 
Specifically, the algorithm iterates through all possible X and Y position and yaw angle samples within the current region calculated by the anytime process. 
Within each iteration, the objects in the first map are transformed using the sampled X, Y, and yaw, and matched against the objects in the second map. 
The algorithm then evaluates the matching score, \todo{which is the total sum of all valid object matches between the two maps.} 
Similar to the data association in \cref{subsec:MS-SLAM Backend}, a valid object model match must have the same semantic label, and their centroid distance and model difference are within the given threshold. 
At the end of each iteration, if the matching score exceeds the current best matching score, it will be recorded along with the valid object matches. 
After all iterations, the best matching score will be compared with the valid loop closure threshold to determine whether a valid loop closure has been identified.

\subsubsection{SlideGraph}
\label{subsec: slidegraph}
The SlideGraph algorithm, unlike the exhaustive search-based SlideMatch algorithm, adopts a descriptor-based matching approach and combines it with a robust outlier rejection method, CLIPPER \cite{lusk2021clipper}, for efficient and robust loop closures. 
While the CLIPPER framework can achieve robust data association by formulating the problem as a maximum clique search in a weighted graph, when an initial data association is unavailable, CLIPPER can still operate under an all-to-all association assumption. However, this quickly becomes too memory demanding and computationally expensive as the number of object landmarks increases in the map. 

To address these limitations, we propose a method for generating initial data associations using descriptor matching. 
Inspired by prior research \cite{nardari2020place}, our descriptor design applies Delaunay triangulation to two sets of object landmarks derived from metric-semantic maps. Candidate simplices, either triangles (2D landmarks) or tetrahedrons (3D landmarks), from each set are compared based on their sorted vertex-to-centroid distances. 
A valid match is found if the distance metrics between two simplices fall within a predefined threshold. 
From these matched simplices, the corresponding vertices are extracted as an initial set of object landmark associations. 
The semantic class labels of the object landmarks are used to further reject false candidate matches. 
This initial matching step significantly reduces the search space for the CLIPPER framework, which subsequently refines these matches by filtering out false associations through its graph-theoretic optimization. 
\todo{The total number of valid matches resulting from this refinement is treated as the matching score, which is then compared against a threshold to determine whether a valid loop closure has been detected. }
This two-stage approach ensures robust data association while maintaining computational feasibility, even for large landmark sets. 

\subsubsection{Least square optimization}
 Finally, a least-squares optimization is applied to obtain the relative transformation estimates between the reference frames of the two metric-semantic maps, using the data associations provided by either the SlideMatch or SlideGraph algorithm. 
Note that we only use estimates for X, Y, Z, and yaw. This is because roll and pitch are directly observable from the IMU, and their estimates from VIO or LIO drift negligibly over time. 
This optimization aims to minimize the Euclidean distance between pairs of matched landmarks found in the data association step.

\begin{table*}[ht!]
\centering
\caption{\textit{Robot platform details.} Specifications on three types of robot platforms used in our experiments.}
\resizebox{0.7\textwidth}{!}{
\begin{tabular}{||c|c|c|c|c||}
\hline
               & Compute             & Primary Sensor        & Battery Life & Autonomous Navigation Speed \\ \hline
               \hline
Falcon 4 UAV   & Intel i7-10710U CPU & Ouster OS1-64         & 30 mins      & 3-10 m/s        \\ \hline
Falcon 250 UAV & Intel i7-10710U CPU & Intel Realsense D435i & 7 mins       & 2-5 m/s         \\ \hline
Scarab UGV     & Intel i7-8700K CPU  & Intel Realsense D455  & 45 mins      & 0.5 m/s         \\ \hline
\end{tabular}
}
\vspace{-0.1in}
\label{table: platfroms}
\end{table*}

\subsection{Decentralized graph optimization for multi-robot SLAM} 
\label{subsec:decentralized graph optimization for multi-robot slam and communication database}

As overviewed in \cref{sec:Decentralized Metric-Semantic SLAM formulation}, for multi-robot scenarios, each robot will exchange information with other robots to find inter-robot loop closures and perform decentralized factor graph optimization. 
When communication is established, the robots will share the lightweight observations stored in their databases, which include object models detected by each key pose and the relative odometry between each pair of key poses. 
Besides, the robots will also share a metric-semantic map containing all object models so that inter-robot loop closure can be effectively performed. 
Messages received from other robots are saved in the database before an inter-robot loop closure is found. 
Once the transformation between the reference frames of two robots is found, all the previously received key poses and object detections from other robots will be transformed into the host robot's reference frame. 
In this way, we can add both the pose nodes and the nodes for detected object models associated with them from other robots to the host robot's factor graph in a similar way as described in \cref{subsec:MS-SLAM Backend}. This process is illustrated in \cref{fig:system-diagram}. The communication and decentralized graph optimization module is illustrated in \cref{fig:decentralized-slam-collaboration-module}. 
Although each robot needs to perform a pose graph optimization using information from all other robots, the induced burden on computation and communication bandwidth is still small due to the sparsity and memory-efficient representation of object models.

One natural question that may arise is why we share such lightweight observations between robots instead of sharing marginalized factor graphs (e.g., as in \cite{cunningham2010ddf, cunningham2013ddf2})? We justify our design choice by the fact that the object-level metric-semantic map that we use is significantly sparser than traditional geometric representations such as point, line, or planar features. In addition, the number of objects detected at each pose is much smaller compared to geometric features (e.g., corner points). 
In terms of performance, our representation allows us to be memory and computationally efficient, as supported by the experimental results in \cref{table: runtime analysis computation}. 
In terms of statistical consistency, our approach does not suffer from spurious information loss due to linearization.

\section{Results and Analysis}
\label{sec:results and analysis}

\subsection{Robot platform overview}
\label{sec:system overview}

\begin{figure}[b!]
        \centering
            \includegraphics[width=0.75\columnwidth]{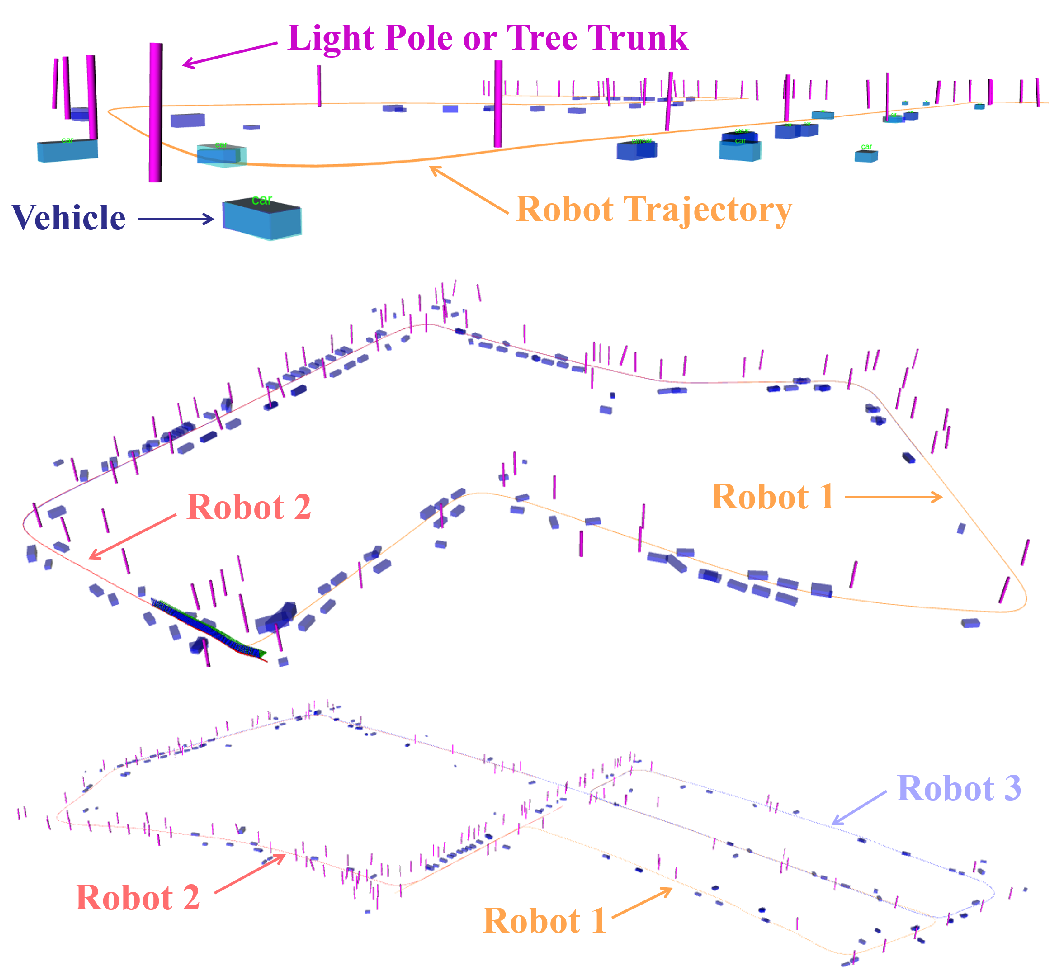}
        \caption{\textit{Metric-semantic SLAM results on the KITTI dataset.} The top, middle, and bottom panels show the results of experiments involving one, two, and three robots, respectively. Each cuboid represents a vehicle while each cylinder represents either a tree trunk or a light pole. The estimated robot trajectories are shown in orange, red, and blue for the first, second, and third robots, respectively.}
        \label{fig:kitti experiment}
\end{figure}

Our robot team consists of three types of aerial and ground robots: the Falcon 250 UAV, the Falcon 4 UAV, and the Scarab UGV, as illustrated in \cref{fig:title-page}. The specifications of these robots are provided in \cref{table: platfroms}.

The Falcon 250 UAV, first introduced in \cite{yuezhantao2023seer} and later improved in \cite{tao20243d}, is a lightweight vision-driven aerial platform capable of autonomous exploration and navigation in cluttered, multi-floor indoor environments. It carries an Intel Realsense D435i RGBD camera, which is the primary sensor for metric-semantic SLAM and autonomous exploration. It is equipped with a VOXL board, which is used for VIO. 
%A Pixhawk 4 Mini flight controller is responsible for low-level attitude control. 
%The total weight of the platform is 1.29 kg and it is powered by a 331 g 3S 5200 milliampere-hour (mAh) lithium polymer battery.

The Falcon 4 UAV \cite{liu2022large} is equipped with a 3D LiDAR, a VectorNav VN-100 IMU, and a Pixhawk 4 flight controller. The platform has % a total weight of 4.2 kg and is powered with a 17,000 mAh lithium-ion battery, achieving 
a flight time of $\sim$30 minutes with all onboard sensors and computer running. 

The Scarab UGV \cite{michael2008experimental} platform is a ground wheeled robot carrying an Intel Realsense D455 camera, which is the primary sensor for metric-semantic SLAM and autonomous exploration. It is also equipped with a 2D Hokuyo LIDAR for low-level odometry and obstacle avoidance. 

All modules in our software system, including instance segmentation, metric-semantic SLAM, exploration, planning, and control, as shown in \cref{fig:system-diagram}, execute in real time onboard the robots.

\subsection{Deployment for autonomous exploration}
\label{sec:integration-with-autonomy}
We integrate the metric-semantic SLAM algorithm with our autonomous exploration and navigation systems for all three types of robot platforms. 
 The integration process and the use of semantic maps differ for different sensing modalities. 
 In \cite{liu2023active}, an active metric-semantic SLAM system with similar map representations is proposed for LiDAR-based UAVs. 
In contrast, the RGB-D-based UAV and UGV platforms in \cite{tao20243d} face larger odometry drifts. To address this, they use the outputs of the metric-semantic SLAM framework to not only explore frontiers to gather more information but also to actively establish semantic loop closures so that the state estimation uncertainty can be reduced.  
 We extend \cite{tao20243d} by (a) upgrading the SLAM module with the metric-semantic SLAM framework proposed in this work so that multi-robot collaboration can be achieved, (b) deploying the active metric-semantic SLAM system that was originally designed for Falcon 250 UAV on a UGV platform (i.e., the Scarab UGV platform), and (c) supporting a heterogeneous team of aerial and ground robots to share information and collaboratively construct metric-semantic maps. %An illustration of metric-semantic SLAM-in-the-loop autonomous exploration can be found in \cref{fig:Visualization of semantic-in-the-loop autonomy}.

\subsection{Experiment design}
\label{subsec: experiment overview}

% \begin{figure}[b!]
%         \centering
%             \includegraphics[trim=0 0 0 400, clip, width=0.8\columnwidth]{img/final-high-res-figures/indoor-robot-team-fig-low-res.png}
%         \caption{\textit{Indoor multi-robot experiment setup.} The robot team involved in indoor semantics-in-the-loop autonomous exploration experiments consists of three Scarab UGVs and one Falcon 250 UAV. The robots start with unknown relative transformations and rely on observing a common set of objects to estimate these transformations and merge their maps in real time.}
%         \label{fig:indoor-multi-robot-pictures}
% \end{figure}

We designed and conducted a set of experiments to evaluate our framework and its modules. We conducted four sets of experiments. 
%A summary of these experiments is provided in \cref{table:four-sets-of-experiments}. 
Here, we provide an overview of each experiment:

    \subsubsection{Experiment 1} 
    \label{subsubsec: experiment 1}
    This experiment is designed to evaluate our decentralized metric-semantic SLAM framework in both indoor and outdoor environments. We utilize the RGBD-equipped Falcon 250 UAV and Scarab UGV, along with the LiDAR-equipped Falcon 4 UAV, each operated by hand-carrying, autonomous navigation, and manual piloting, respectively. These experiments demonstrate the versatility of our metric-semantic SLAM system across heterogeneous platforms and environments. The experimental setup is divided into two distinct sub-experiments.

    \textit{(1.a)} 
    The Falcon 4 UAV is manually piloted across three different parking lots, while the Falcon 250 is hand-carried through four sequential runs inside two buildings. Each of the seven runs starts with a pre-defined pose. To test the capabilities of our semantics-driven place recognition and loop closure algorithms, we create unknown initial poses for each run by streaming the data from different starting timestamps. During data playback, the sensor data from each robot is processed on a base station with an i7-10750H CPU (comparable to CPUs onboard the robots), where each robot operates on a separate ROS node. We enable communication between the ROS nodes of different robots at predetermined intervals, effectively replicating an intermittent and opportunistic communication paradigm. This setup allows us to evaluate the robustness and efficacy of our decentralized metric-semantic SLAM framework across heterogeneous robots and environments.
        
     \textit{(1.b)} This sub-experiment begins with the Falcon 4 UAV and the Scarab UGV inside a building. The Falcon 4, hand-carried, maps the entire first floor, while the Scarab autonomously navigates and maps the same area. The initial poses of the robots are unknown, posing a challenge in place recognition and map merging. This experiment demonstrates the effectiveness of our semantics-driven place recognition and map merging module in handling data from heterogeneous sensing modalities, including LiDARs and RGBD cameras.

    \subsubsection{Experiment 2} 
    \label{subsubsec: experiment 2}
    A multi-robot simultaneous indoor autonomous exploration experiment where the Falcon 250 and Scarab platforms actively explore the building and jointly construct a metric-semantic map. All robots in the experiment explore the environment fully autonomously and share information through either constantly communicating with each other or communicating with the base station at the end of the mission.
    The robots leverage the metric-semantic map in real time for their active SLAM systems, which trades off between exploration and uncertainty reduction. 
    This experiment can be further broken down into two sub-experiments.

    \textit{(2.a)} The Falcon 250 and Scarab platforms start at different locations on the first floor of a building. They then autonomously explore the environment to the best of their abilities and build a metric-semantic map. At the end of the mission, they share their information to the base station in an attempt to merge them.  The transformation between the initial starting positions of the two robots is unknown. This experiment demonstrates the ability of our system to operate on a heterogeneous team of robots and be efficient and lightweight enough to leave enough computational headroom to support their autonomy stack.

    \textit{(2.b)} Three Scarab platforms are tasked with the operation of autonomously exploring the first floor of a building. They build a metric-semantic map in a decentralized manner and share information through constant communication using Wi-Fi in an attempt to construct a merged map. In addition to demonstrating efficient operations on resource-constrained robots, this experiment also highlights our multi-robot decentralized metric-semantic SLAM module and how it processes sensor information to facilitate online collaborative SLAM with inter-robot loop closures and map merging.

    \subsubsection{Experiment 3} 
    \label{subsubsec: experiment 3}
    An experiment in forest environments where the Falcon 4 is manually piloted to different sections of a forest for a total of three sequential runs to map all the tree trunks. The Falcon 4 platform starts from the same position for each run (for ground truthing purposes) but the data is processed in a manner that mimics a three-robot simultaneous operation with an unknown initial transformation between the robots. This was achieved by processing the three bags of data and streaming them from different timestamps such that it replicates the situation of three robots starting at various locations with a sufficient translation between them. Forest environments are a steep contrast to urban environments as they are more cluttered with dense and homogeneous object landmarks (i.e. tree trunks). 
    By experimenting in such environments, we showcase our large-scale mapping efficiency and our ability to merge maps from multiple robots in dense and perceptually aliased environments.     
    In addition, obtaining information on semantic objects in forests, such as tree trunks and branches, can enhance silviculture practices, quantifying carbon sequestration, and combating climate change.

    \subsubsection{Experiment 4} 
    \label{subsubsec: experiment 4}
    In the final experiment, we tested the proposed framework on the publicly available Semantic KITTI dataset. This allowed us to demonstrate the algorithm's versatility and effectiveness in scenarios beyond our custom robot platforms and experimental setups.

\begin{table*}[ht!]
\caption{\textit{Inter-robot localization errors.} 
SM and SG refer to the SlideMatch and SlideGraph algorithms, respectively. Position error is calculated as the L2 norm of the errors in the X, Y, and Z coordinates of the relative transformation between two robots' reference frames. In general, the SlideGraph algorithm is more efficient and requires less parameter tuning compared to SlideMatch, while still providing comparable results. However, both algorithms encounter difficulties in scenarios where object landmarks are noisy, particularly with RGBD-based sensing. SlideGraph is relatively more prone to failures in such conditions due to insufficient pairwise consistency in the positions of object landmarks.
%For SlideMatch, the average position error across all experiments is 0.220 m with a standard deviation of 0.148 m. The average yaw error is -0.16$\deg$ with a standard deviation of 2.48$\deg$.
}
 \setlength%\extrarowheight{5pt}
 \resizebox{1.0 \textwidth}{!}
 {\begin{tabular}{||c | c | c | c | c ||} 
 \hline
 Experiment (Sensor) & Position Err. (SM) [m] & Yaw Err. (SM) [$\deg$] & Position Err. (SG) [m] & Yaw Err. (SG) [$\deg$] \\
 \hline
  {Outdoor-Urban} Falcon 4 \#0 to Falcon 4 \#1 (LiDAR) &  0.235  & -0.2  & 0.074 & 0.3 \\%-0.00291  \\
    \hline
  {Outdoor-Urban} Falcon 4 \#1 to Falcon 4 \#2 (LiDAR)  & 0.159 & -0.3  & 0.240 & 0.9 \\ %-0.00467   \\
    \hline
  {Outdoor-Urban} Falcon 4 \#0 to Falcon 4 \#2 (LiDAR)  & 0.135 & -0.9  & 0.848 & 0.7 \\%-0.0150\\
    \hline
 {Outdoor-Forest} Falcon 4 \#0 to Falcon 4 \#2 (LiDAR)  & 0.040 & 3.1  & 0.101 & 0.3 \\%0.0547 \\
    \hline
 {Outdoor-Forest} Falcon 4 \#1 to Falcon 4 \#2 (LiDAR) & 0.159 & 3.2  & 0.049 & -3.0 \\% 0.0563   \\
    \hline
 {Outdoor-Forest} Falcon 4 \#0 to Falcon 4 \#1 (LiDAR) & 0.054 & -0.8  & 0.126 & -2.6 \\% -0.0135   \\
    \hline
 {Cross Sensing Modality} (LiDAR-RGBD) & 0.542  & -4.0  & 0.578 & 2.0 \\% -0.0698  \\
  \hline
 {Indoor} Scarab 45 to Scarab 40 (RGBD) & 0.221 & -3.4  & 0.952 & -2.3 \\% 0.0593  \\
  \hline
 {Indoor} Scarab 40 to Scarab 41 (RGBD) & 0.165 & 0.0  & 0.253 & 0.1 \\% 0.0002 \\
  \hline
 {Indoor} Scarab 45 to Scarab 41 (RGBD) & 0.460 & 3.7  & 1.040 & -6.1 \\%0.0637 \\
 \hline
 {Indoor Aerial-Ground} (RGBD) & 0.245 & -2.4 & -- & -- \\ %-0.0411  \\
  \hline
\end{tabular}}
\label{table: Inter-robot localization and map merging quantitative results}
\end{table*}

\subsection{Indoor-outdoor LiDAR-RGBD metric-semantic SLAM without inter-robot loop closures}
\label{subsubsec: Qualitative indoor-outdoor with known TF 7 robots}

\begin{figure}[t!]
        \centering
            \includegraphics[width=0.75 \columnwidth]{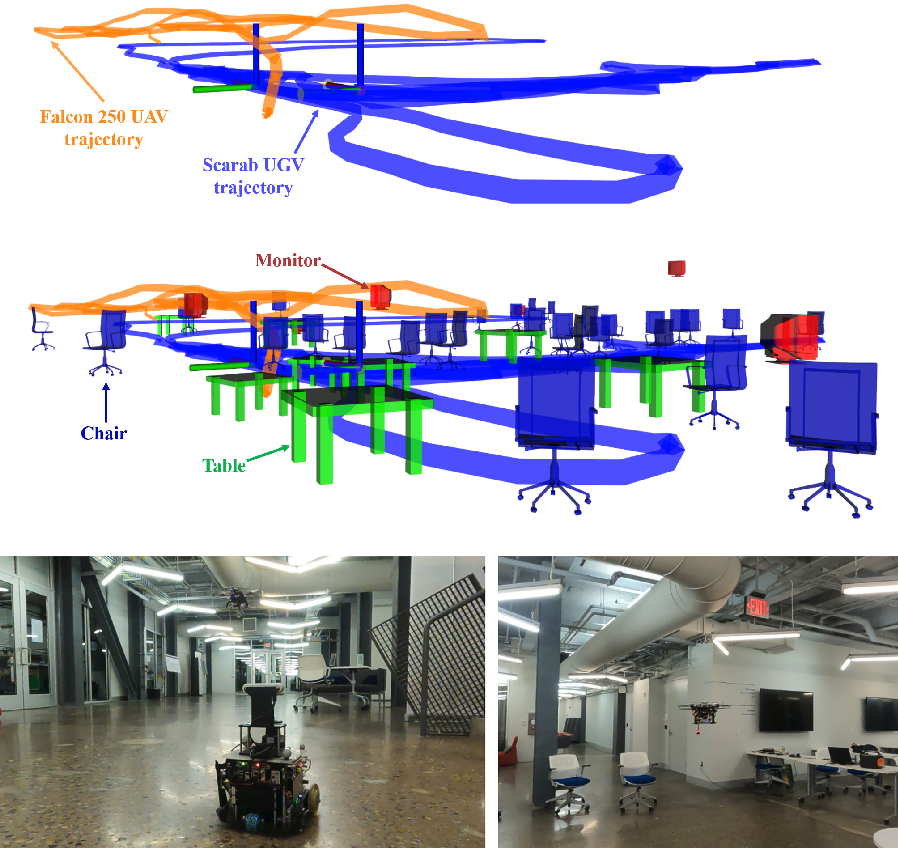}

        %     %%%%%%%%%%%%%%%%%%%%%%  HIGH RESOLUTION %%%%%%%%%%%%%%%%%%%%%%
        
        %     \includegraphics[trim=100 900 100 500, clip, width=0.9\columnwidth]{img/f250-scarab-autonomy/trajectory-f250-scarab-oblique-view.png}
        % \centering
        %     \includegraphics[trim=650 410 100 800, clip, width=0.9\columnwidth]{img/f250-scarab-autonomy/trajectory-view-f250-scarab-zoomed.png}
        % \centering
        %     \includegraphics[trim=670 150 120 750, clip, width=0.9\columnwidth]{img/f250-scarab-autonomy/oblique-map-view-f250-scarab-zoomed.png}
            
        %         \centering
        %     \includegraphics[trim=100 0 0 0, clip, height=1.1in]{img/f250-scarab-autonomy/1-four-robots-f250-scarab.png}
        %     \includegraphics[trim=200 0 200 0, clip, height=1.1in]{img/f250-scarab-autonomy/show-monitor-four-robots-f250-scarab.png}
        %     %%%%%%%%%%%%%%%%%%%%%%  HIGH RESOLUTION %%%%%%%%%%%%%%%%%%%%%%

        \caption{\textit{Exploration and metric-semantic SLAM with heterogeneous robot teams.} {This figure shows the trajectories and metric-semantic map constructed from autonomous exploration experiments by aerial and ground robots. The Falcon 250 UAV explores in 3D (orange trajectory), while the Scarab UGV explores in 2D (blue trajectory). Our method can merge maps across UAVs and UGVs by leveraging the viewpoint invariance of semantic landmarks. 
        % Note that some of the monitors in the environment are hanging on the wall (as shown in the upper right panel), thus appearing to be higher than the table and chairs.
        }}
        \label{fig:f250-scarab-map-merging-autonomy-experiment}
\end{figure}

\cref{fig:f250-scarab-map-merging-autonomy-experiment} shows qualitative results on metric-semantic SLAM for experiment 1.a as described in \cref{subsec: experiment overview}. 
Outdoor objects such as cars, tree trunks, and light poles and indoor objects such as chairs, tables, and monitors are detected and mapped using our sparse metric-semantic map representation. 
Object models for vehicles are represented by blue cuboids, while tree trunks and light pole landmarks are shown as pink cylinders. 
\todo{The ellipsoid models of chairs, tables, and monitors are replaced with corresponding Computer-Aided Design (CAD) models. This is done only for visualization purposes and an intuitive interpretation of the metric-semantic map, especially in highly cluttered indoor environments.} These models are colored blue, green, and red, respectively, according to their detected semantic class. This experiment shows the capabilities of our metric-semantic SLAM framework in integrating data from multiple robots equipped with either LiDARs or RGBD cameras, enabling the construction of a unified metric-semantic map that includes both indoor and outdoor areas. 

\subsection{Urban and forest LiDAR-only metric-semantic SLAM with inter-robot loop closures}
\label{subsubsec: semantic place recognition inter-robot loop closure results lidar}
\label{subsubsec: Qualitative outdoor with unknown TF urban forest}

\begin{figure}[t!]
        \centering
            \includegraphics[width=0.75\columnwidth]{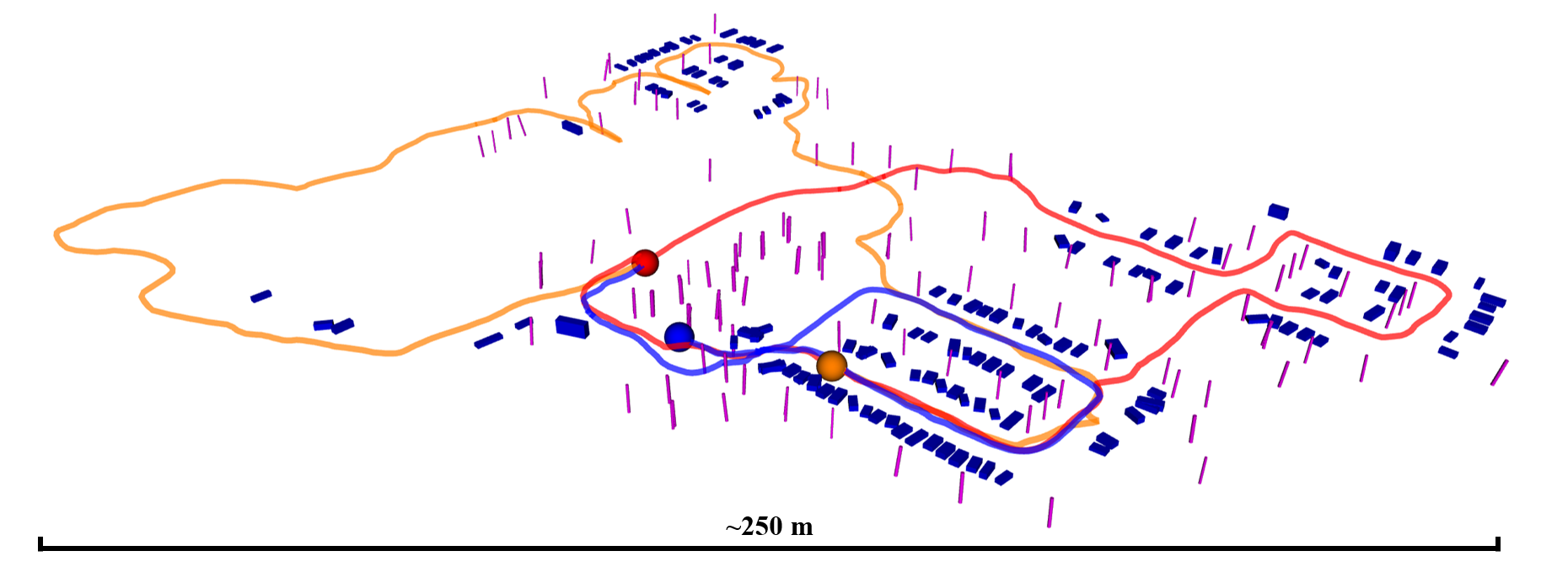}
        \caption{\textit{Multi-robot metric-semantic SLAM in an outdoor experiment.} Results from an experiment involving three Falcon 4 UAV flights. The trajectories, marked by red, blue, and yellow bold lines, start from different positions (red, blue, and yellow dots) but converge back to the same end location (red dot). Our semantics-driven place recognition algorithm was able to detect inter-robot loop closures, which are then used to merge the maps. In the map, vehicle models are represented by blue cuboids, and models for tree trunks or light poles are shown by magenta cylinders.}
        \label{fig:outdoor experiment}
\end{figure}
\begin{figure}[t!]

        \centering
        \includegraphics[trim=0 0 30 0, clip, width=0.4\textwidth]{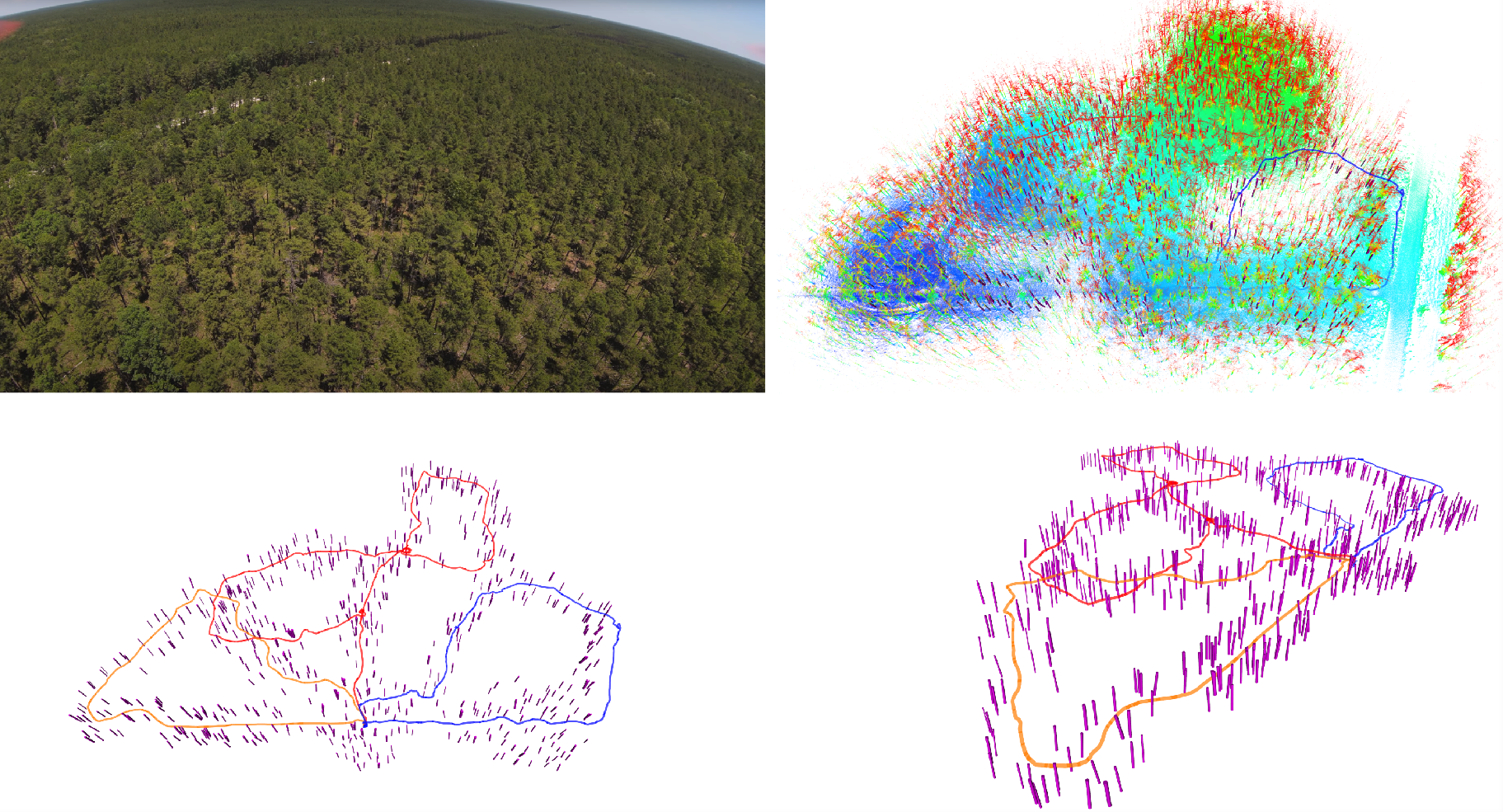}
        % \includegraphics[trim=0 0 0 0, clip, width=1.0\textwidth]{img/compressed-figures/new-compressed-forest-three-robot.png}

        % %%%%%%%%%%%%%%%%%%%%%%%%%%%%%%%%% HIGH RESOLUTION %%%%%%%%%%%%%%%%%%%%%%%%%%%%%%%%%%%%%%%%%%%%%%%
        % \centering
        %         \includegraphics[trim=0 0 0 0, clip, width=0.99\columnwidth]{img/forest-wharton-forest-overhead.png}
        %         \includegraphics[trim=0 0 0 0, clip, width=0.99\columnwidth]{img/forest-experiment/Screenshot from 2024-02-06 15-14-32.png}
        
        % \centering          
        %         \includegraphics[trim=0 0 0 0, clip, width=0.99\columnwidth]{img/forest-experiment/Screenshot from 2024-02-06 15-14-53.png}
        %         \includegraphics[trim=0 0 0 0, clip, width=0.99\columnwidth]{img/forest-experiment/Screenshot from 2024-02-06 15-15-38.png}
        % %%%%%%%%%%%%%%%%%%%%%%%%%%%%%%%%% HIGH RESOLUTION %%%%%%%%%%%%%%%%%%%%%%%%%%%%%%%%%%%%%%%%%%%%%%%
        
        \caption{\textit{Forest experiments.} This figure illustrates the overhead UAV view of the environment (top left), metric-semantic map with point clouds (top right), and object-based map constructed by three under-canopy UAV flights (bottom row). Three robots with their trajectories colored in red, yellow, and blue can merge their information to the starting location with unknown initial transformations between their starting poses. The figure shows metric-semantic maps of tree trunks (modeled as cylinders) overlayed on top of an aggregated point cloud showing the forest environment. The trajectory of each robot, in this case, is $\sim$1 km in length, and 860 trees have been mapped. The mapped trees span an area of $\sim$160,000 $m^2$.
        % Qualitative analysis is provided in \cref{subsubsec: Qualitative outdoor with unknown TF urban forest}, and quantitative results are provided in \cref{subsubsec: semantic place recognition inter-robot loop closure results lidar}.
        }
        \label{fig:forest-maps}

\end{figure}

\cref{fig:outdoor experiment} shows qualitative results on multi-robot metric-semantic SLAM, demonstrating inter-robot loop closure and map merging capabilities in outdoor environments using LiDAR-equipped Falcon 4 UAVs. This experiment utilizes data from experiment 1.a. In this specific setup, the transformation between the robots is unknown and thus has to be estimated to merge their maps.  
 \cref{fig:forest-maps} shows qualitative results for the forest experiments, with the overhead view illustrating that the environment is large-scale, unstructured, and features dense homogeneous objects (i.e. trees). 
In both experiments, our semantics-driven place recognition algorithm reliably detects inter-robot loop closures while accurately estimating relative transformations between the robots.

\begin{figure}[t!]       
\centering
\includegraphics[width=0.64\columnwidth]{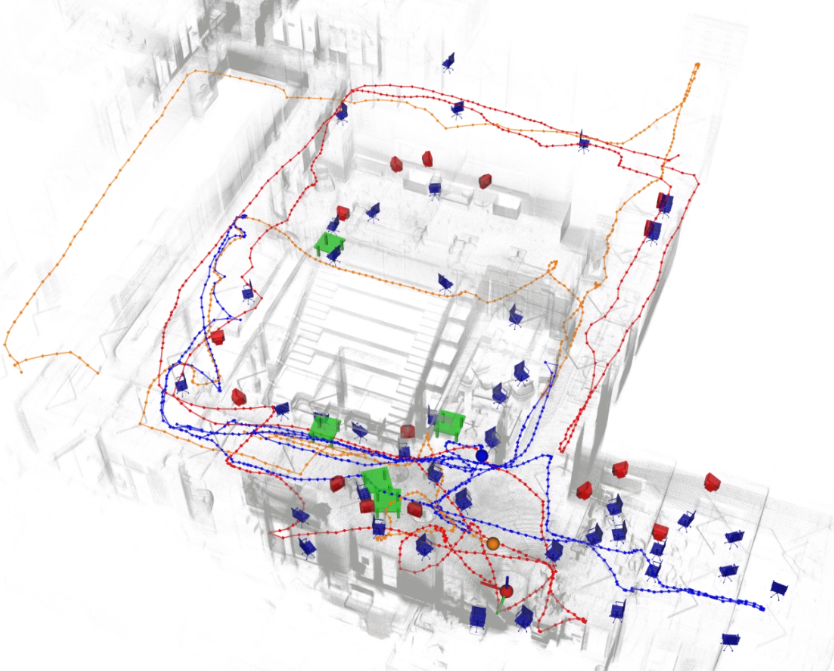}

        \caption{\textit{Metric semantic map constructed in real time onboard three Scarab UGVs.} The UGVs autonomously explore the environment while communicating with each other. Blue, orange and red trajectories correspond to three trajectories from the three UGVs. Our semantic place recognition algorithm was able to perform inter-robot loop closure, which is used to construct a merged metric-semantic map. The gray-colored accumulated point cloud is not generated by these UGVs but by a LiDAR robot for visualization purposes. 
        The figure overlays the trajectories with the metric-semantic map to qualitatively assess the validity of the results.
        }
        \label{fig:scarab_with_comm three robot indoor map merging}
\end{figure}

The quantitative results on inter-robot localization are shown in \cref{table: Inter-robot localization and map merging quantitative results}. In urban environments, our algorithm utilizes semantic object models of various classes and shapes to estimate the relative 3D position and yaw between each pair of robots. In forest settings, our method leverages models of tree trunks for inter-robot localization. 
The SlideMatch algorithm achieves an average position error of 0.13 m and a yaw error of 0.68$\deg$, with standard deviations of 0.07 m and 1.76$\deg$, respectively.
In contrast, the SlideGraph algorithm shows an average position error of 0.24 m and a yaw error of -0.57$\deg$, with standard deviations of 0.28 m and 1.60$\deg$, respectively. 

Despite significant variations in object classes, appearance, and density when transitioning from urban to forest environments, both algorithms consistently demonstrate robust and accurate performance. Furthermore, the SlideGraph algorithm requires less parameter tuning and, unlike SlideMatch which relies on a computationally expensive exhaustive search, is more efficient.

\subsection{Indoor RGBD-only aerial-ground metric-semantic SLAM with inter-robot loop closures}
\label{subsubsec: semantic place recognition inter-robot loop closure results rgbd}
\label{subsubsec: Qualitative indoor-outdoor with unknown TF}
\label{subsubsec: semantic place recognition results aerial ground}

% \cref{fig:f250-scarab-map-merging-autonomy-experiment} illustrates the experimental setup (i.e., experiment 2 in \cref{subsec: experiment overview}) showing the indoor environment and the robot team consisting of three UGVs and one UAV. 
% \cref{fig:Visualization of semantic-in-the-loop autonomy} presents an internal representation of the Falcon 250 UAV, which includes the voxel and metric-semantic maps along with the long-horizon exploration path and dynamically feasible local trajectories, all constructed in real time while the UAV actively explores the environment. 
% The representation of the Scarab UGV is largely similar to this, except that the exploration path and local trajectories are constrained to 2D space.
% \cref{fig:f250-scarab-map-merging-autonomy-experiment} shows an example of the indoor environment and the robot team consisting of aerial and ground robots. 
In this experiment, the starting position and orientation of the robots vary significantly, with up to 6 m in position and a 90$\deg$ difference in yaw. \cref{fig:scarab_with_comm three robot indoor map merging} shows the results constructed by merging sensor measurements from three UGVs autonomously exploring the environment.   
Our semantics-driven place recognition algorithm was able to detect and accurately estimate relative transformations as robots accumulate observations. 
Based on the estimated relative transformations, our metric-semantic factor graph fuses the information from multiple robots to construct a merged metric-semantic map.

The quantitative results are shown in \cref{table: Inter-robot localization and map merging quantitative results}. For the indoor experiment with multiple Scarab UGVs, our SlideMatch algorithm achieves an average position error of 0.28 m and a yaw error of 0.10°, with standard deviations of 0.13 m and 2.90°, respectively. Our SlideGraph algorithm achieves an average position error of 0.75 m and a yaw error of -2.77°, with standard deviations of 0.35 m and 2.55°, respectively.

\cref{fig:f250-scarab-map-merging-autonomy-experiment} shows the results constructed by the Falcon 250 and one of the UGVs that autonomously explore the environment. The invariance of semantic object models to viewpoint changes enables our algorithm to merge measurements from different viewpoints as observed by heterogeneous aerial and ground robots. 
As shown in the last row of \cref{table: Inter-robot localization and map merging quantitative results}, the SlideMatch algorithm successfully identifies inter-robot loop closures and estimates the relative transformation with a position error of 0.245 m and a yaw error of 2.35$\deg$. However, we observed that aerial-ground localization exhibits slightly larger errors along the Z-axis compared to localization errors among identical robot platforms. This discrepancy arises because aerial robots primarily observe the upper parts of objects, while ground robots focus on the lower parts, leading to greater variation in Z-axis estimates. 
In contrast, the SlideGraph algorithm fails to perform inter-robot loop closures between RGBD-based aerial and ground robots. This limitation is caused by reduced pairwise consistency between object positions in the map, which stems from noisy RGB-D sensing combined with the fact that the aerial and ground robots observe different parts of the objects.

In summary, while noisy RGBD sensing can introduce inaccuracies in the localization and modeling of individual objects, a collection of object landmarks provides the robots with an informative description of the environment. This enables them to establish loop closures even under drastic viewpoint changes. In addition, it is important to note that the increased noise levels require a careful selection of parameters, such as the search discretization in SlideMatch or the descriptor matching threshold in SlideGraph.

\begin{figure}[t!]
    \centering          
    \includegraphics[trim=120 10 0 0, clip, width=0.5\columnwidth]{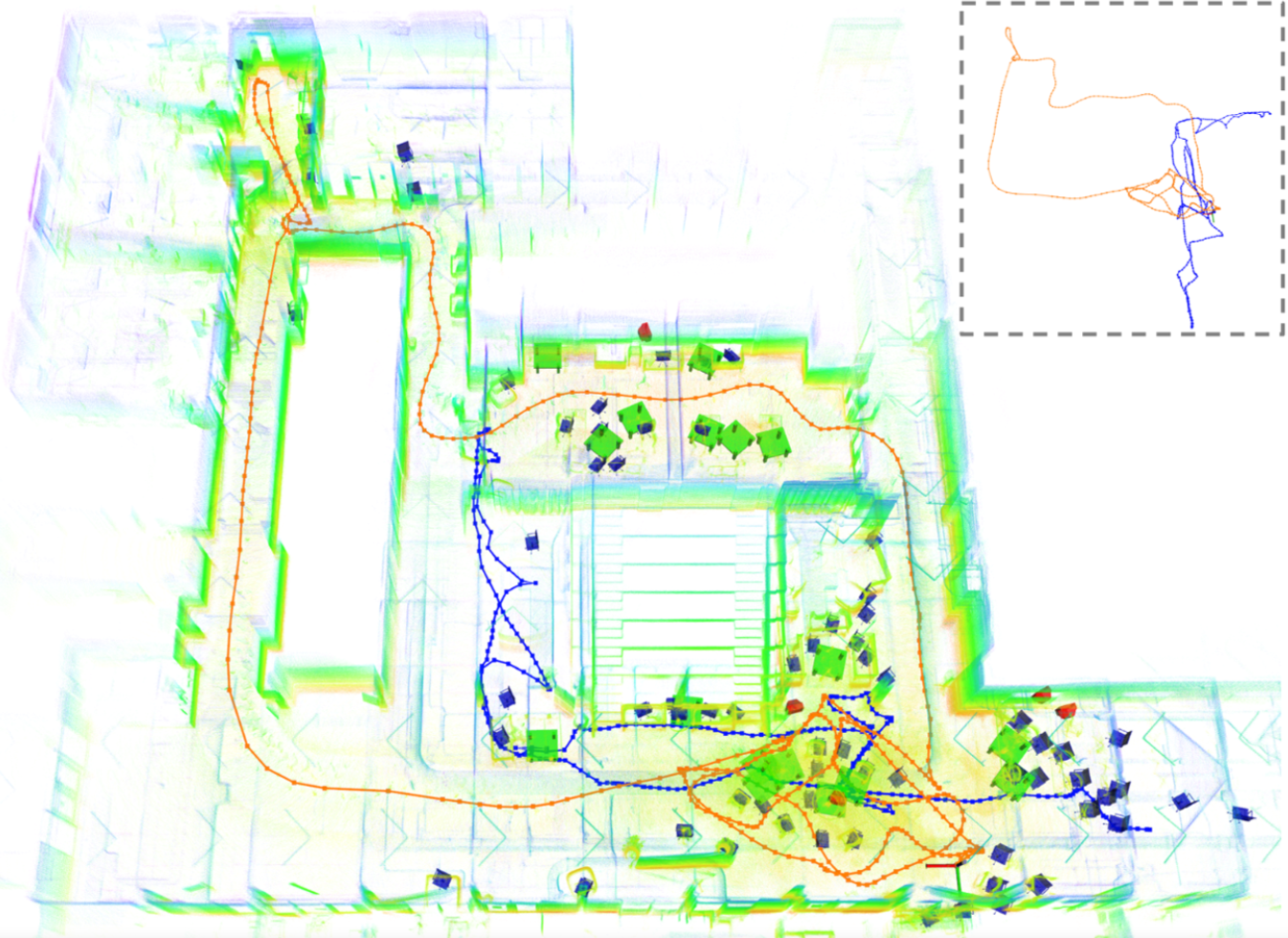}
        \caption{\textit{Cross sensing modality place recognition.}
        A LiDAR-equipped robot is represented by the orange trajectory and an RGBD-based robot by the blue trajectory. The rest of the panels show the merged metric-semantic map constructed by the robots and overlayed on top of the accumulated LiDAR point cloud. The upper-right panel illustrates the robots' trajectories in their own reference frames. This comparison shows how the place-recognition and loop-closure algorithm effectively registers the two robots with different sensing modalities into a common reference frame despite the drastic differences in raw sensor data and the initial poses.}
        \label{fig:Cross sensing modality place recognition}
\end{figure}

\subsection{Indoor LiDAR-RGBD metric-semantic SLAM with inter-robot loop closures}

\label{subsubsec: semantic place recognition results cross sensing modality}
\label{subsubsec: Qualitative cross sensing modality with unknown TF}
\cref{fig:Cross sensing modality place recognition} shows qualitative results on multi-robot metric-semantic SLAM on the cross sensing modality experiment (i.e., experiment 1.b in \cref{subsec: experiment overview}). 
The accumulated point cloud captured by the LiDAR robot is overlaid on the semantic object models mapped by both the LiDAR and the RGBD robots. 
Note that there is a relative transformation of 2.51 m along X, -5.37 m along Y, and 90$\deg$ along yaw between the reference frames of the two robots. 
Our algorithm leverages the invariance of semantic object models across sensing modalities. This is a unique advantage of our method given that geometric features extracted from RGB or depth images significantly differ from those derived from LiDAR point clouds. This capability enables accurate estimation of the transformation between robots equipped with different sensors—specifically \textit{LiDAR} and \textit{RGBD}—allowing us to fuse their measurements and create a merged metric-semantic map.

Quantitative results are shown in \cref{table: Inter-robot localization and map merging quantitative results}. Our semantics-driven place recognition algorithms demonstrate the ability to establish loop closures between robots equipped with different sensors. SlideMatch estimates the relative transformation with a position error of 0.542 m and a yaw error of -4.00$\deg$, while SlideGraph results in a position error of 0.578 m and a yaw error of 2.00$\deg$. The accuracy of inter-robot localization in cross-sensing-modality scenarios is slightly lower compared to single-modality experiments, particularly those involving LiDAR-based robots. This performance decline can be attributed to differences in fields of view, object detection accuracy, and noise levels between LiDAR and RGBD sensors. These factors lead to greater discrepancies in object modeling, which, in turn, reduce the precision of inter-robot localization.

\subsection{Object mapping quantitative results}
\label{subsec: object mapping results}
\label{subsubsec: Object mapping quantitative results}
We report precision, recall, and F1 scores on object mapping results of both cars and tree trunks across multiple experiments. 
We first obtain the ground truth by surveying the environment and manually marking down the positions of objects. 
Here, we focus the object mapping results on cars and tree trunks since ground truth for them can be reliably acquired when compared to other objects such as chairs, tables or monitors.
This is because cars are usually parked in clearly separated parking spaces and tree trunks are often spaced apart at regular intervals.
A True Positive (TP), or a valid object match, occurs when an object in our metric-semantic map is confirmed to exist at a position that aligns with the ground truth.  
A False Positive (FP) refers to an object that appears in our metric-semantic map but is not present in the ground truth. 
Conversely, a False Negative (FN) refers to an object that is present in the ground truth but does not appear in our metric-semantic map. 
These values are then used to calculate precision, recall, and F1 scores. 
The results can be found in \cref{table:Quantitative results on object mapping.}.  Note that those results are acquired using the smallest backbone (darknet-13 as detailed in 
\cite{liu2023active}) for the semantic segmentation model, which is less accurate but can run inference in real time using the onboard NUC computer. The results can be further improved using larger backbones such as darknet-26 or darknet-53, which are supported in \cite{milioto2019rangenet++}. 
As shown in \cref{table:Quantitative results on object mapping.}, the average precision for cars is 0.967 and for tree trunks is 0.946. The average recall for cars is 0.881 and for tree trunks is 0.909. The F1 score is 0.922 for cars and 0.927 for tree trunks. 
In summary, through these metrics, we demonstrate that our method is able to reliably detect and map objects that exist in the environment, with very few false positives or false negatives.
This result is important for downstream tasks such as inventory estimation.

\begin{table}[b!]
\begin{center} 

\caption{\textit{Quantitative results of object detection: Ground Truth (G.T.) vs. Estimated (Est.).} Precision is evaluated by determining the proportion of detected objects that accurately match the ground truth, thus measuring the algorithm’s object mapping accuracy. Recall indicates the likelihood of the algorithm successfully detecting an object present in the environment, reflecting the algorithm’s object mapping sensitivity.}
 \setlength
 \resizebox{0.42\textwidth}{!}
 {\begin{tabular}{||c | c | c | c | c | c ||} 
 \hline
 Envir. & G.T. Cars & Est. Cars & Precision & Recall & F1  \\
 \hline
 Lot 1 & 42 & 35 & 1.00 & 0.833 & 0.909  \\
  \hline
  Lot 2  & 61 & 59 & 0.932 & 0.902 & 0.917  \\
  \hline
 Lot 3 & 32 & 29 & 1.00 & 0.906 & 0.951 \\
  \hline
 Total & 135 & 123 & 0.967 & 0.881 & 0.922 \\
 \hline
 \hline
 Envir. & G.T. Trees & Est. Trees & Precision & Recall & F1  \\
 \hline
 Total & 77 & 74 & 0.946 & 0.909 & 0.927 \\
    \hline
\end{tabular}}
\end{center}

\label{table:Quantitative results on object mapping.}
\end{table}

\subsection{Communication usage and runtime analysis}
\label{subsec:  runtime analysis computation}

\subsubsection{Discussion on scalability}

For the factor graph optimization, suppose we have $G$ variables for each object landmark and $B$ total object landmarks. The total number of variables for all landmarks is $S = G \cdot B$. 
When adding a new factor, the number of variables that need to be solved for is $G$ for a new object factor, and $9$ for a new odometry (i.e., $\mathbb{SE}$(3) relative pose) factor in the factor graph optimization process \cite{kaess2012isam2}. 
G is a small number, and specifically in our formulation, the variables corresponding to each cuboid, cylinder, and ellipsoid semantic landmark node are 9, 9, and 5, respectively.
Therefore, this complexity becomes $\mathcal{O}{(1)}$ for adding any new factor. 
Upon loop closure, the optimization runs in time $\mathcal{O}\left(max(F, S)^{1.5}\right)$ \cite{kaess2012isam2}. 
Practically, in traditional SLAM systems that are based on geometric features, the number of features usually far exceeds the number of robot poses. As a result, $max(F, S) = S$, which is usually at the order of $100F$ (i.e. each key frame tracks 100 features) or even larger. 
Due to the use of sparse semantic landmarks in the proposed system, $S$ will usually be of the order $\mathcal{O}{(F)}$. Therefore, to sum up, the complexity of factor graph optimization is $\mathcal{O}{(1)}$ during normal operation and $\mathcal{O}{(F^{1.5})}$ during loop closures. 

When running with multi-robot teams, the framework runs in a decentralized manner; therefore, for each robot, the complexity scales linearly with the size of the robot team (that is, $K$). As a result, the complexity of decentralized factor graph optimization is $\mathcal{O}{(K)}$ when adding new factors coming from the robot team, and $\mathcal{O}{((K \cdot F)^{1.5})}$ during loop closures. In practice, the size of the robot team $K$ is much smaller than the number of poses for each robot $F$; therefore, the decentralized factor graph optimization can be executed in real-time onboard the robot.

\subsubsection{Experimental results}

In this section, we show the computational and communication usage when running our metric-semantic SLAM framework. 
All results are obtained using a computer equipped with an Intel i7-10750H CPU, without the support of a GPU. 
This CPU's performance is comparable to that of the onboard computers in the robots.
In \cref{table: runtime analysis computation}, the average and maximum size of communication packets are summarized when robots operate with intermittent communication. 
To replicate intermittent communication, each robot publishes ROS messages to each other every 10 seconds. 
Benefiting from the sparse metric-semantic map representation, our algorithm is shown to be communication efficient when deployed on real robots. 
The similarity in communication usage across different robots stems from the fact that we replicate all robots as being within communication range by streaming their data on different ROS nodes running on a laptop. 
For indoor experiments, communication is facilitated via Wi-Fi. 
Although each robot shares all the information in its database with neighboring robots within the communication range, we can further reduce the communication burden by implementing a bookmarking system that only publishes new information since the last communication between robots.

\begin{table*}[]

\caption{\textit{Runtime and communication results.} 
%The number of objects is the count of object landmarks in our metric-semantic map. 
The runtime for factor addition and graph optimization accounts for handling both the observation of new landmarks and loop closure scenarios. The runtime of semantics-driven place recognition and loop-closure algorithms includes both loop closure detection and relative transformation estimation. Note that the place-recognition and loop-closure process runs on a separate thread. Therefore, its latency does not impact real-time operation. SlideGraph requires significantly lower computational runtime, particularly in environments with fewer object landmarks. This efficiency is attributed to the use of descriptor-based matching, where computational demands are influenced not by the scale of the environment but only by the number of object landmarks. 
In contrast, SlideMatch exhaustively searches across the entire search region, and its runtime depends on various factors, such as the search discretization. 
The communication usage is measured by the size of messages published by each robot. Here, the average usage is calculated by dividing the total size of all messages by the number of messages, while the maximum size corresponds to the size of the last messages since they usually contain the most observations.}
\resizebox{\textwidth}{!}{
\begin{tabular}{|c|c|c|c|cc|cc|}
\hline
\multirow{2}{*}{} & 
\multirow{2}{*}{\begin{tabular}[c]{@{}c@{}}Number of \\ Objects\end{tabular}} & 
\multirow{2}{*}{Robot ID} & 
\multirow{2}{*}{\begin{tabular}[c]{@{}c@{}}Trajectory \\ Length {[}m{]}\end{tabular}} & 
\multicolumn{2}{c|}{Runtime {[}ms{]}} & 
\multicolumn{2}{c|}{Communication Usage {[}MB{]}} \\ \cline{5-8} 
 & & & & 
\multicolumn{1}{c|}{\begin{tabular}[c]{@{}c@{}}{Factor Adding and} \\ {Graph Optimization}\end{tabular}} & 
\begin{tabular}[c]{@{}c@{}} {Inter-Robot Loop Closure} \\ {SlideGraph / SlideMatch}\end{tabular} & 
\multicolumn{1}{c|}{\begin{tabular}[c]{@{}c@{}}{Average}\\ {Communication Usage}\end{tabular}} & 
\begin{tabular}[c]{@{}c@{}}Maximal \\ Communication Usage\end{tabular} \\ \hline

\multirow{3}{*}{Indoor} & 
\multirow{3}{*}{68} & 
0 & 
172.072 & 
\multicolumn{1}{c|}{{2.429}} & 
 {22} / 3638 & 
\multicolumn{1}{c|}{{0.304}} & 
0.578 \\ \cline{3-8} 
 & & 
1 & 
155.930 & 
\multicolumn{1}{c|}{{4.008}} & 
{24} / 1614 & 
\multicolumn{1}{c|}{{0.304}} & 
0.575 \\ \cline{3-8} 
 & & 
2 & 
194.625 & 
\multicolumn{1}{c|}{{2.372}} & 
{28} / 2252 & 
\multicolumn{1}{c|}{{0.309}} & 
0.576 \\ \hline

\multirow{3}{*}{Outdoor Urban} & 
\multirow{3}{*}{249} & 
0 & 
699.543 & 
\multicolumn{1}{c|}{{9.351}} & 
{95} / 131 & 
\multicolumn{1}{c|}{{0.458}} & 
0.725 \\ \cline{3-8} 
 & & 
1 & 
264.264 & 
\multicolumn{1}{c|}{{7.763}} & 
{86} / 123 & 
\multicolumn{1}{c|}{{0.453}} & 
0.727 \\ \cline{3-8} 
 & & 
2 & 
410.194 & 
\multicolumn{1}{c|}{{13.296}} & 
{91} / 168  & 
\multicolumn{1}{c|}{{0.466}} & 
0.725 \\ \hline

\multirow{3}{*}{Outdoor Forest} & 
\multirow{3}{*}{860} & 
0 & 
326.476 & 
\multicolumn{1}{c|}{{28.702}} & 
{209} / 1124 & 
\multicolumn{1}{c|}{{1.702}} & 
2.686 \\ \cline{3-8} 
 & & 
1 & 
367.798 & 
\multicolumn{1}{c|}{{28.638}} & 
{406} / 2091 & 
\multicolumn{1}{c|}{{1.735}} & 
2.687 \\ \cline{3-8} 
 & & 
2 & 
746.870 & 
\multicolumn{1}{c|}{{52.046}} & 
{375} / 1836 & 
\multicolumn{1}{c|}{{1.742}} & 
2.686 \\ \hline
\end{tabular}%
}
\label{table: runtime analysis computation}
\end{table*}

Additionally, the runtime for factor adding and graph optimization, and loop closure (if applicable), is reported in \cref{table: runtime analysis computation}.
Except for the loop closure module, which is used to identify inter-robot loop closures and operates on a separate CPU thread, all other modules are triggered each time new observations arrive. 

\todo{The front-end runtime depends on the object segmentation and detection model.
Our results show that on a \textit{single thread} of Intel i7-10750H CPU, the Rangenet++ \cite{milioto2019rangenet++} front-end with our custom darknet-13 model runs at 1.5 hertz (Hz). 
The front-end with the YOLOv8 \cite{Jocher_YOLOv8_by_Ultralytics_2023}, specifically, YOLOv8m, runs at 1 Hz per CPU thread. 
When multiple CPU threads are used, the inference speed can be increased. For example, YOLOv8m can run at 3 Hz using 4 out of 12 CPU threads. The cuboid, cylinder, and ellipsoid modeling process runs at the corresponding object detection rate.}

\todo{The back-end factor graph adding and graph optimization processes show different patterns across different environments. For the indoor environment, the average runtime of graph optimization is 2$\sim$4 ms, with 68 landmarks.} 
%with an average time per landmark of 0.036 ms. 
The outdoor urban environment exhibits an average runtime of graph optimization of 10.14 ms, with 249 landmarks. The average time per landmark in this setting is 0.041 ms. Due to the increased density of landmarks in the outdoor forest environment, the average runtime spikes to 36.46 ms, with a significantly higher number of landmarks at 860.
%, and an average time per landmark of 0.042 ms. 

These results illustrate that the average runtime required for factor graph adding and graph optimization increases with the number of landmarks. This relationship is influenced by the presence of inter-robot loop closures in these experiments, which makes the computational complexity lie between observing new landmarks and processing loop closures. 

The runtime also increases with the trajectory length, which is partly due to more pose nodes being added to the graph. However, the correlation between trajectory length and runtime is not as strong as the correlation between the number of landmarks and runtime. An increase in the runtime per meter traveled by the robot is observed as the environment transitions from less complex settings (indoor) to more complex ones (outdoor forest). This trend also shows the challenges and computational costs associated with large-scale environments. 

Although the runtime of the semantics-driven place recognition and loop closure algorithms is longer than that of factor graph optimization, the resulting latency is acceptable. In addition, since the semantics-driven place recognition and loop closure algorithm operate in a separate CPU thread, this latency does not directly impact real-time localization and navigation. Once the computation is complete, the estimated loop closure information can be used by the factor graph to update both the map and the robot trajectories.

In addition to the importance of \textit{sparse semantic} landmarks in facilitating long-range autonomous exploration and navigation, our results on communication provide another reason for using semantic landmarks in multi-robot collaboration tasks. 
The average communication per landmark is 3.96 kilobytes (KB) for indoor environments, 1.84 KB for outdoor environments, and 2.03 KB for forest environments. 
The average communication bandwidth per meter traveled by the robot is 1.65 KB in indoor settings, 1.85 KB in outdoor settings, and 5.13 KB in forest settings. 
As the density of landmarks in the environment increases, communication overhead also increases. 
However, the overall communication bandwidth required remains minimal.
This modest communication demand is particularly important when robots operate under intermittent and bandwidth-limited communication conditions. 
In addition, this efficiency in communication also shows the effectiveness of \textit{sparse semantic} mapping in minimizing resource usage. 
%These properties provide our framework with the potential to integrate with advanced multi-robot communication frameworks \cite{cladera2023enabling} or resource-efficient multi-robot coordination methods \cite{xu-Tzoumas-Vasileios2024communication}. 

To sum up, the results show that, although there is some variance across different environments due to the difference in density of objects, the use of \textit{sparse semantic} map representations offers the robot a rich understanding of its surroundings while maintaining efficiency in computation, storage, and communication. Our proposed method not only facilitates real-time execution of exploration and navigation tasks but also enables multi-robot collaboration across various environments on a large scale.

\subsection{Analysis of SlideMatch and SlideGraph algorithms}
\label{sec:discussion}
\label{sec:limitations-and-future-work}

% \subsection{SlideMatch vs. SlideGraph} 
\label{subsec:slidematch vs. slidegraph}
Throughout these experiments, we have observed unique advantages, limitations, and deployment considerations for both the SlideGraph and SlideMatch algorithms. SlideGraph demonstrates superior performance with minimal parameter tuning in scenarios where the object-level metric-semantic map experiences relatively low noise in object position estimates. 
Therefore, SlideGraph is generally the preferred option, particularly for LiDAR-based experiments or scenarios involving large search spaces. 
On the other hand, the SlideMatch algorithm can sometimes achieve successful loop closure in situations where the SlideGraph method fails, particularly in experiments with large errors in object position estimates. 
This is because the pairwise consistency assumption used in \cite{lusk2021clipper} becomes less suitable under such conditions. 
Such situations are common in indoor RGBD-based object mapping, where cluttered spaces lead to small differences in object-to-object distances and high noise in position estimates due to noisy depth measurements, noisy local odometry, limited field of view, and complex object shapes. 
Although semantic labels can help disambiguate these measurements to some extent, it remains challenging to use pairwise consistency to distinguish outliers from inliers. 
In contrast, SlideMatch leverages the entire map and operates similarly to a \textit{voting} method, making it more tolerant to noise in local pairwise object distances. As a result, SlideMatch performs better in such worst-case scenarios, albeit at the cost of significant computational overhead.

\subsection{Benchmarks}

{
In this section, we present a series of benchmark experiments evaluating our system’s performance in terms of localization accuracy, computational efficiency, and memory usage. 
We report results from (1) evaluating our framework on standard public datasets and (2) comparing it with other open-source methods. 
We note that benchmarking the entire system is challenging due to the differences in components across other metric-semantic SLAM systems.
Therefore, we opt to isolate and benchmark key modules instead. This allows for a fair and focused comparison of our system. 
}

\begin{table*}[ht]
\caption{
{\textit{Benchmark comparison between SlideSLAM (SlideGraph) and existing methods.} For these benchmarks, we report results between robot 0 and robot 1, as referenced by the Semantic KITTI experiments in \cref{subsec: semantic kitti results} and forest experiments in \cref{subsubsec: Qualitative outdoor with unknown TF urban forest}. Note that the map sizes for both CLIPPER and SlideSLAM are the same, as they use the same object maps as inputs. Also, note that for the memory and runtime benchmarks of LCDNet, the results reported are the memory consumed for each individual point cloud and the runtime to process a pair of point clouds. In practice, it scales linearly with the number of point clouds. Cells filled with a ``-" indicate that the algorithm failed.}  
}
\resizebox{\textwidth}{!}{
\begin{tabular}{|c|ccc|ccc|ccc|ccc|}
\hline
      Metric   & \multicolumn{3}{c|}{{Trans. Err (m)}}                               & \multicolumn{3}{c|}{{Abs. Yaw. Err (degree)}     }                          & \multicolumn{3}{c|}{{Map / Input Size}}                                                    & \multicolumn{3}{c|}{{Runtime (ms)}}                                    \\ \hline
       Method  & \multicolumn{1}{c|}{CLIPPER}   & \multicolumn{1}{c|}{LCDNet} & Ours      & \multicolumn{1}{c|}{CLIPPER}    & \multicolumn{1}{c|}{LCDNet} & Ours       & \multicolumn{1}{c|}{CLIPPER (full maps)}        & \multicolumn{1}{c|}{LCDNet (per scan)} & Ours (full maps)             & \multicolumn{1}{c|}{CLIPPER  (CPU)} & \multicolumn{1}{c|}{LCDNet (per scan pair on GPU)} & Ours (CPU)    \\ \hline
{KITTI 05} & \multicolumn{1}{c|}{0.04} & \multicolumn{1}{c|}{0.09}       & 0.03 & \multicolumn{1}{c|}{0.04}  & \multicolumn{1}{c|}{0.22}       & 0.08 & \multicolumn{1}{c|}{7.2 KB} & \multicolumn{1}{c|}{0.84 MB}       & 7.2 KB & \multicolumn{1}{c|}{152} & \multicolumn{1}{c|}{1580}       & 40 \\ \hline
{KITTI 07} & \multicolumn{1}{c|}{0.06} & \multicolumn{1}{c|}{0.05}       & 0.09 & \multicolumn{1}{c|}{0.01} & \multicolumn{1}{c|}{0.21}       & 0.09  & \multicolumn{1}{c|}{7.3 KB}  & \multicolumn{1}{c|}{0.84 MB}       & 7.3 KB & \multicolumn{1}{c|}{202} & \multicolumn{1}{c|}{1580}       & 50 \\ \hline
{Forest}   & \multicolumn{1}{c|}{-}         & \multicolumn{1}{c|}{1.10}       & 0.13 & \multicolumn{1}{c|}{-}          & \multicolumn{1}{c|}{0.1}       & 2.6 & \multicolumn{1}{c|}{16 KB}    & \multicolumn{1}{c|}{0.84 MB}       & 16 KB & \multicolumn{1}{c|}{-}       & \multicolumn{1}{c|}{1580}       & 209 \\ \hline

\end{tabular}
   }
\label{table: SOTA benchmarks}
\end{table*}

\subsubsection{Bechmark on Semantic KITTI}
\label{subsec: semantic kitti results}
In addition to conducting experiments with our own robots, we also tested our algorithm framework on the publicly available Semantic KITTI dataset. 
The framework takes in the point clouds, initial pose estimation from SuMa++ \cite{chen2019suma++}, and ground truth segmentation, and estimates the metric-semantic map as well as the robot trajectory. 

The qualitative results, as shown in \cref{fig:kitti experiment}, demonstrate the ability of the proposed framework to construct metric-semantic maps and merge measurements from multiple robots. 
We conducted a quantitative evaluation of our SlideGraph algorithm for inter-robot loop closure using sequences $\#5$ and $\#7$. Sequence $\#7$ was divided into three intervals, i.e., (0, 600), (400, 1000), and (500, 1100), for the first, second, and third robots, respectively. Sequence $\#5$ was similarly divided into three intervals, i.e., (0, 1000), (700, 2000), and (1600, 2760), to create datasets for each of the three robots.
Our algorithm successfully identified inter-robot loop closures and accurately estimated the relative transformations between the robots. For sequence $\#5$, the SlideGraph algorithm results in a position error of 0.03 m and 0.05 m, and a rotation error of 0.080$\deg$ and 0.065$\deg$ for the transformations between robot 0 and robot 1, and between robot 1 and robot 2, respectively. In sequence $\#7$, the algorithm results in a position error of 0.093 m and 0.095 m, and a rotation error of 0.096$\deg$ and 0.097$\deg$. 
\todo{For sequence \#5 and sequence \#7, the total trajectory lengths of three robots are  2.8 km and 1.2 km. The absolute trajectory errors of the trajectories of all three robots are 0.998 m \lz{(vs. 1.154 m from SuMa++)} and 0.962 m \lz{(vs. 1.148 m from SuMa++)}, respectively. The relative translation errors are 0.111\% and 0.255\%. The relative rotation errors are 0.0011 degrees/m and 0.00256 degrees/m.} 

These results not only demonstrate the accuracy of inter-robot localization but also illustrate the effectiveness of the proposed algorithm framework when deployed on public datasets, showing its potential as a plug-and-play tool for multi-robot metric-semantic SLAM.

\subsubsection{Benchmark against existing loop closure methods} 
{
To provide a comprehensive comparison, we benchmark SlideSLAM against two existing approaches from the literature.
These include: (1) {LCDNet}~\cite{cattaneo2022lcdnet}, a deep-learning-based loop closure method for LiDAR data; and (2) {CLIPPER}~\cite{lusk2021clipper}, a graph-theoretic algorithm for data association.
}

{
LCDNet performs scan-to-scan place recognition using a learned 3D feature extractor and registration network. We benchmark the algorithm using weights that are pretrained on the KITTI dataset. We use the same sequences and splits of Semantic KITTI as described in the previous benchmark. We also benchmark LCDNet on our forest data, as described in the previous sections. The benchmarks for LCDNet were conducted on a GPU-equipped desktop (with an RTX 3070 GPU). 
The inter-robot localization results are shown in \cref{table: SOTA benchmarks}. The results are comparable in the KITTI environment. However, for the forest environment, SlideSLAM provides better localization results. This shows the advantage of our method to generalize to different environments as long as the objects are detected. Our method requires no additional fine‑tuning data for the loop closure step, whereas LCDNet does.  Moreover, we also benchmark the memory requirements to store the observations or maps required for the loop closure step. For LCDNet, these representations can be individual processed point cloud scans (0.84 MB), point features extracted from the PV-RCNN network for each point cloud scan (4.18 MB), or the final global descriptors of those features for each point cloud scan (10.0 MB). As a result, enabling inter-robot loop closure detection with LCDNet requires a minimum bandwidth of 0.84 MB per LiDAR scan. In contrast, as shown in the table, the entire object map of SlideSLAM requires less than 10 KB for the KITTI sequences, which contain hundreds of LiDAR scans. The sparsity of our map representation offers a key advantage for large-scale multi-robot experiments, as it significantly reduces both computational and communication requirements compared to methods operating on less compact map representations.}

{
{CLIPPER}~\cite{lusk2021clipper} serves as the data association backbone of our SlideSLAM (SlideGraph) module. However, the original CLIPPER lacks semantic information and relies on exhaustive pairwise matching over all object pairs, resulting in high computational demands in complex scenes (e.g., each of our forest maps consists of hundreds of landmarks). Our SlideGraph enhances CLIPPER by using semantic class labels and polygon-descriptor-based initial data association, enabling us to significantly reduce the hypothesis space. CLIPPER is benchmarked on similar sequences on Semantic KITTI as detailed in the previous section, and also on our forest dataset. The results are shown in \cref{table: SOTA benchmarks}. In challenging environments such as forests, the original CLIPPER algorithm encountered memory overflows due to the large number of landmarks, which significantly expanded the hypothesis space. 
Take the forest environment as an example: a pair of maps with approximately 300 and 800 landmarks would generate around 240,000 hypotheses. In contrast, our SlideGraph algorithm prunes this raw hypothesis space down to approximately 3,000, significantly reducing the size of the consistency graph by 2 orders of magnitude. 
As a result, SlideGraph consistently achieved robust performance with lower memory usage and shorter runtimes. 
}

\section{Discussion on Limitations and Future Work}
In this section, we address a few limitations of the current system and how they motivate our future work. Firstly, the proposed sparse map representation is limited to three simple shapes for modeling objects. While it helps maintain sparsity, the map may not always be visually informative with these shapes, and they may fail to model the finer metric details for objects with complex shapes, which are needed for downstream tasks such as robot manipulation. 
{In addition, they may lead to information loss that results in inaccuracies or failures to identify loop closures, especially in environments with ambiguous or repetitive object layouts. Semantic aliasing may also occur when different objects are represented by the same primitive, increasing dependence on accurate semantic labeling. Misclassifications in such cases can cause errors.}
We are interested in investigating better map representations that can model generic objects in detail while still maintaining relative sparsity. Secondly, the current inter-robot loop closure strategy does not follow a conventional ``filtering" approach, where multiple loop closures are first established and then a consensus is formed to choose the best loop closure result and filter out false positives. Currently, we only use a sufficient inlier threshold to filter out false positives and establish loop closure in a ``one-shot" manner. This may not always be effective in certain situations where the environment is cluttered with the same semantic landmarks and similar pairwise distances (e.g., two conference rooms with similar table and chair arrangements, or dense forests). Therefore, we would like to employ a filtering-based approach to our current loop closure strategy, where the best loop closure candidate is chosen based on a maximum consensus of potential loop closure options. Lastly, our current place recognition and mapping formulation doesn't account for dynamic objects. Such objects (e.g., a car changing its parking spot over time) violate the pairwise distance consistency assumption required for robust place recognition and can therefore induce faulty loop closures. In the future, we would like to leverage semantic knowledge of objects, combined with intelligent agents such as Large Language Models (LLMs), to better handle dynamic objects. Using the semantic knowledge, an LLM can infer which objects can be dynamic in nature and automatically filter them out during place recognition. 

\section{Conclusion}
\label{sec:conclusion}

In this work, we introduce a real-time decentralized metric-semantic SLAM framework, along with its integration into autonomous exploration and navigation systems for aerial and ground robots. 
This integrated system enables teams of heterogeneous robots to autonomously explore and collaboratively construct maps with both geometric and open-vocabulary semantic information across a variety of indoor and outdoor environments. 
The robots opportunistically leverage communication to exchange sparse, lightweight semantic measurements, which are used for inter-robot localization and map merging. 
Through a comprehensive set of real-world experiments involving three types of aerial and ground robots, \revise{as well as benchmarks on publicly available datasets and comparisons against existing methods}, we demonstrate the capabilities, effectiveness, and robustness of our system. 
Our runtime and communication analysis also show the importance of using sparse semantic map representations, especially for large-scale (over 1 km) metric-semantic mapping tasks in challenging and dense environments, such as forests with over 1000 object landmarks.      
\section{Acknowledgment}
We thank Jie Mei for his help with experiments, Guilherme V. Nardari, Ian D. Miller, Kashish Garg, Jeremy Wang, and Alex Zhou for the software and hardware support.

\bibliographystyle{IEEEtran}
\bibliography{ref, ag-survey-ref, phd_qual_ref}

\section{Author's Biography Section}

\vspace{-0.5 in}

% \bf{If you include a photo:}\vspace{-283pt}
% \begin{IEEEbiographynophoto}
\begin{IEEEbiography}[{\includegraphics[width=1in,height=1.25in, trim=50 0 50 0, clip,keepaspectratio]{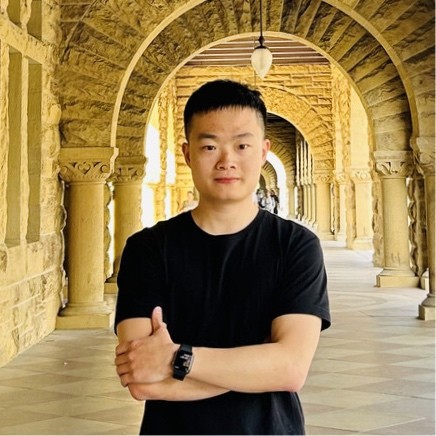}}]{Xu Liu} received the M.S.E. degree in Robotics and the Ph.D. degree in Mechanical Engineering and Applied Mechanics from the General Robotics, Automation, Sensing, and Perception (GRASP) Laboratory, University of Pennsylvania, Philadelphia, PA, USA, in 2019 and 2024, respectively. He was a Postdoctoral Scholar at Stanford University, Stanford, CA, USA. He is currently a Senior Applied Scientist at Microsoft, Redmond, WA, USA. His research interests include metric-semantic SLAM, autonomous UAVs, and foundation models for robotics.
\end{IEEEbiography}

\vspace{-0.5 in}

\begin{IEEEbiography}[{\includegraphics[width=1in,height=1.25in,clip,keepaspectratio]{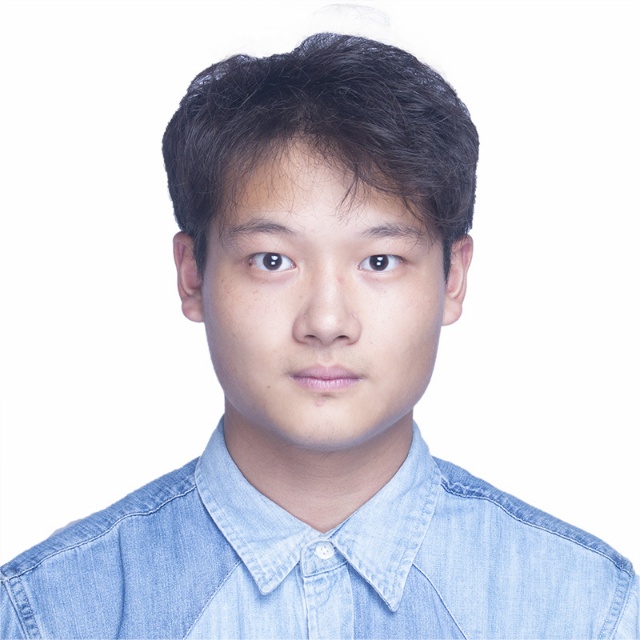}}]
% \begin{IEEEbiographynophoto}
{Jiuzhou Lei} received his bachelor's degree in Mechanical Engineering from Sichuan University, Chengdu, China in 2021 and the M.S.E. degree in Robotics from the GRASP Laboratory, University of Pennsylvania, Philadelphia, PA, USA in 2023. He is currently a Ph.D. student in Mechanical Engineering at Texas A\&M University, College Station, Texas, USA. He works on robot autonomy, general-purpose robotic manipulation, and navigation.
\end{IEEEbiography}

\vspace{-0.4 in}

\begin{IEEEbiography}[{\includegraphics[width=1in,height=1.25in,clip,keepaspectratio]{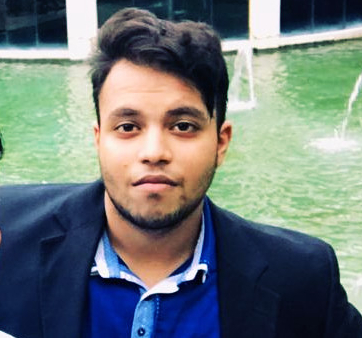}}]
% \begin{IEEEbiographynophoto}
{Ankit Prabhu} received the B.Tech. degree in Computer Science and Engineering from SRM University, Chennai, India, in 2021, and the M.S.E. degree in Robotics from the GRASP Laboratory, University of Pennsylvania, Philadelphia, PA, USA, in 2023. He is currently a Research Scientist at the KumarRobotics lab, and his research focuses on metric-semantic SLAM and robot autonomy. 
\end{IEEEbiography}

\vspace{-0.4 in}

\begin{IEEEbiography}[{\includegraphics[width=1in,height=1.25in,clip,keepaspectratio]{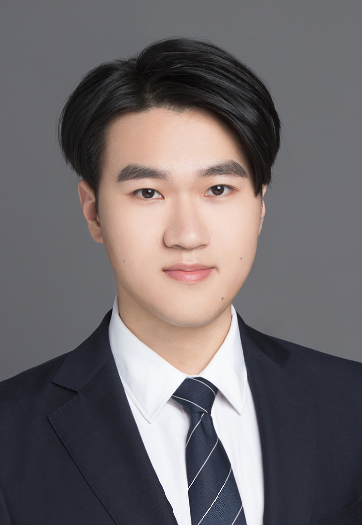}}]
{Yuezhan Tao} received the B.Eng. (Hons.) degree in computer science and engineering from The Chinese University of Hong Kong, Shenzhen, Guangdong, China, in 2019, the M.S.E degree in computer and information science from the University of Pennsylvania, Philadelphia, PA, USA in 2021. He is currently a Ph.D student in computer and information science with the University of Pennsylvania, Philadelphia, PA, USA. His research focuses on active perception, with an emphasis on mapping and motion planning for autonomous robotic systems.
\end{IEEEbiography}

\vspace{-0.4 in}

\begin{IEEEbiography}[{\includegraphics[width=1in,height=1.25in, trim=400 0 400 250, clip,keepaspectratio]{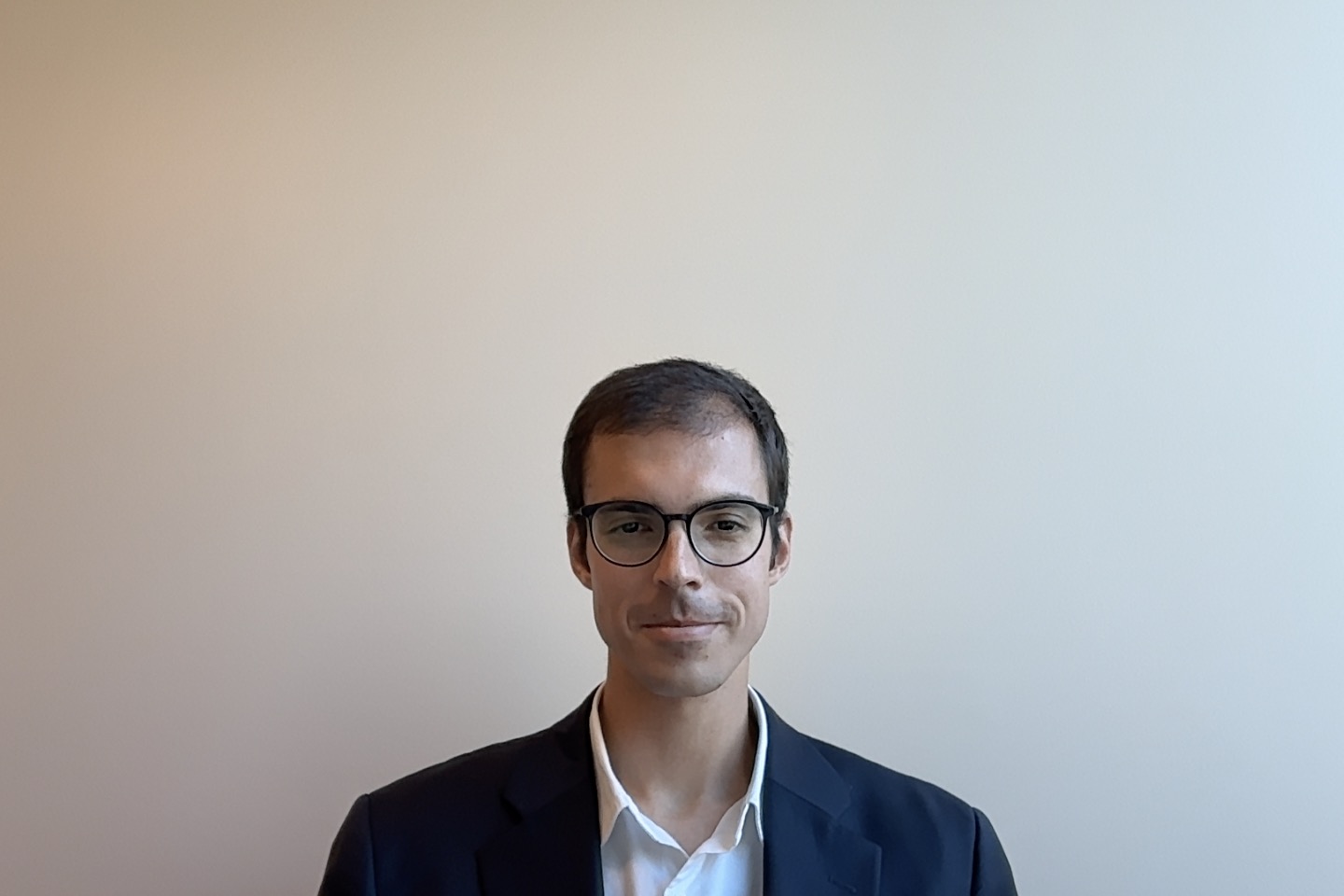}}]
{Igor Spasojevic} is an Assistant Professor at the University of California, Riverside. Before joining UCR, he was a postdoctoral scholar in the GRASP Laboratory at the University of Pennsylvania, Philadelphia, PA. He received his Ph.D. from the Massachusetts Institute of Technology, and his master's and bachelor of arts degrees from the University of Cambridge. His interests focus on optimization algorithms in robotics, including topics such as motion planning, active sensing, and autonomous exploration.
\end{IEEEbiography}

\vspace{-0.4 in}

\begin{IEEEbiography}[{\includegraphics[width=1in,height=1.25in,clip,keepaspectratio]{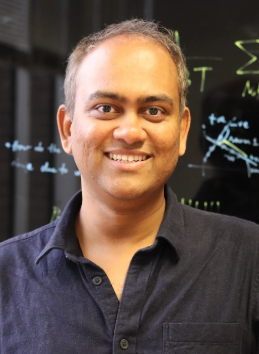}}]
{Pratik Chaudhari} is an Assistant Professor in Electrical and
Systems Engineering and Computer and Information Science at the
University of Pennsylvania. He is a core member of the GRASP
Laboratory. From 2018-19, he was a Senior Applied Scientist at Amazon
Web Services and a Postdoctoral Scholar in Computing and Mathematical
Sciences at the California Institute of Technology, Pasadena, CA, USA.
He received his PhD in Computer Science from
University of California, Los Angeles, CA, USA, in 2018, and his Master's and
Engineer's degrees in Aeronautics and
Astronautics from Massachusetts Institute of Technology, Cambridge, MA, USA, in
2012 and 2014, respectively.
He was a part of NuTonomy Inc. (now Hyundai-Aptiv Motional) from 2014-16.
He is the recipient of the
Amazon Machine Learning Research Award in 2020, National Science Foundation
CAREER award in 2022 and the Intel Rising Star Faculty Award in 2022.

\end{IEEEbiography}

\vspace{-0.4 in}

\begin{IEEEbiography}[{\includegraphics[width=1in,height=1.25in,clip,keepaspectratio]{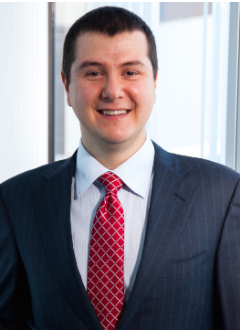}}]
{Nikolay Atanasov}(Senior Member, IEEE) received a B.S. degree in Electrical Engineering from Trinity College, Hartford, CT, USA, in 2008, and M.S. and Ph.D. degrees in Electrical and Systems Engineering from the University of Pennsylvania, Philadelphia, PA, USA in 2012 and 2015, respectively. He is an Associate Professor of Electrical and Computer Engineering with the University of California San Diego, La Jolla, CA, USA. His research focuses on 
% robotics, control theory, and machine learning with emphasis on active perception problems for autonomous mobile robots. He works on 
probabilistic models for simultaneous localization and mapping (SLAM) and on optimal control and reinforcement learning algorithms for minimizing model uncertainty. Dr. Atanasov was the recipient of the Joseph and Rosaline Wolf award for the best Ph.D. dissertation in Electrical and Systems Engineering at the University of Pennsylvania in 2015, the Best Conference Paper Award at the IEEE International Conference on Robotics and Automation (ICRA) in 2017, the NSF CAREER Award in 2021, and the IEEE RAS Early Academic Career Award in Robotics and Automation in 2023.
\end{IEEEbiography}

\vspace{-0.4 in}

\begin{IEEEbiography}[{\includegraphics[width=1in,height=1.25in,clip,keepaspectratio]{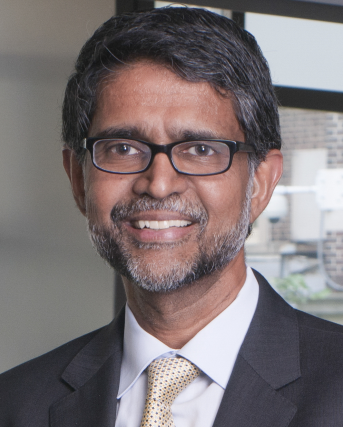}}]
{Vijay Kumar}(Fellow, IEEE) received the Ph.D. degree in mechanical engineering from
the Ohio State University, Columbus, OH, USA, in 1987. He is currently the Nemirovsky
Family Dean of Penn Engineering (with appointments) with the Department of
Mechanical Engineering and Applied Mechanics, the Department of Computer and
Information Science, and the Department of Electrical and Systems Engineering,
University of Pennsylvania, Philadelphia, PA, USA. From 2012 to 2014, he was the
Assistant Director of Robotics and Cyber Physical Systems with the White House Office
of Science and Technology Policy. has served on the editorial boards of the IEEE
Transactions on Robotics and Automation, IEEE Transactions on Automation Science
and Engineering, ASME Journal of Mechanical Design, and the Springer Tract in
Advanced Robotics (STAR), and was the chief editor for the ASME Journal of
Mechanisms and Robotics. He has won best paper awards at DARS 2002, ICRA 2004,
ICRA 2011, RSS 2011, RSS 2013, ICRA 2014, BICT 2015, and MARSS 2016 and has
advised doctoral students who have won Best Student Paper Awards at ICRA 2008,
RSS 2009, and DARS 2010. He was recognized with the RSS Time of Test Award in
2025. Additionally, he is the recipient of the 2013 Popular Mechanics Breakthrough
Award, a 2014 Engelberger Robotics Award, the 2017 IEEE Robotics and Automation
Society George Saridis Leadership Award, the 2018 IEEE Robotics and Automation
Pioneer Award, and the 2020 IEEE Robotics and Automation Field Award. He was
elected to the National Academy of Engineering in 2013, the American Philosophical
Society in 2018, and the American Academy of Arts and Sciences and the National
Academy of Inventors in 2022.
\end{IEEEbiography}


% \bf{If you will not include a photo:}\vspace{-83pt}

\end{document}